\documentclass[manuscript]{acmart}
\usepackage[ruled,vlined,linesnumbered]{algorithm2e}
\DontPrintSemicolon                          
\SetKwInOut{Input}{Inputs}                   
\SetKwInOut{Output}{Output}
\SetKw{KwTo}{to}                             
\SetKw{KwBy}{by}
\usepackage{algpseudocode} 
\usepackage{enumerate}
\usepackage{enumitem}
\usepackage{tabularx}
\usepackage{wrapfig}
\usepackage{placeins}
\usepackage[normalem]{ulem}

\AtBeginDocument{%
  }

\setcopyright{acmlicensed}
\copyrightyear{2018}
\acmYear{2018}
\acmDOI{XXXXXXX.XXXXXXX}
\acmConference[Conference acronym 'XX]{Make sure to enter the correct
  conference title from your rights confirmation email}{June 03--05,
  2018}{Woodstock, NY}
\acmISBN{978-1-4503-XXXX-X/2018/06}

\begin{document}

\title{Interactive Program Generation for Modeling Collaborative Physical Activities from Narrated Demonstrations}

\author{Edward Kim}
\authornote{These authors contributed equally to this research.}
\email{ek65@eecs.berkeley.edu}

\author{Daniel He}
\authornotemark[1]
\author{Jorge Diaz Chao}
\authornotemark[1]

\affiliation{%
  \institution{University of California, Berkeley}
  \country{USA}
}

\author{Wiktor Rajca, Mohammed Amin, Nishant Malpani}
\affiliation{%
  \institution{University of California, Berkeley}
  \country{USA}
}

\author{Ruta Desai}
\affiliation{%
  \institution{Meta}
  \country{USA}
}

\author{Antti Oulasvirta}
\affiliation{%
  \institution{Aalto University}
  \country{Finland}
}

\author{Bjoern Hartmann, Sanjit Seshia}
\affiliation{%
  \institution{University of California, Berkeley}
  \country{USA}
}


\begin{abstract}
Teaching systems physical tasks is a long-standing goal in HCI, yet most prior work has focused on non-collaborative physical activities. Collaborative tasks introduce added complexity, requiring systems to infer users’ assumptions about their teammates’ intent—an inherently ambiguous and dynamic process. This necessitates representations that are interpretable and correctable, enabling users to inspect and refine system behavior. We address this challenge by framing collaborative task learning as a program generation problem. Our system represents behavior as editable programs and uses narrated demonstrations—paired physical actions and natural language—as a unified modality for teaching, inspecting, and correcting system logic without requiring users to see or write code. The same modality is used for the system’s responses, enabling intuitive two-way interaction. In a within-subjects study, 20 users taught multiplayer soccer tactics to our system. Participants successfully refined learned programs to match their intent and found the system easy to use.
\end{abstract}


\keywords{Do, Not, Us, This, Code, Put, the, Correct, Terms, for,
  Your, Paper}
\begin{teaserfigure}
  \centering
  \includegraphics[width=1\textwidth]{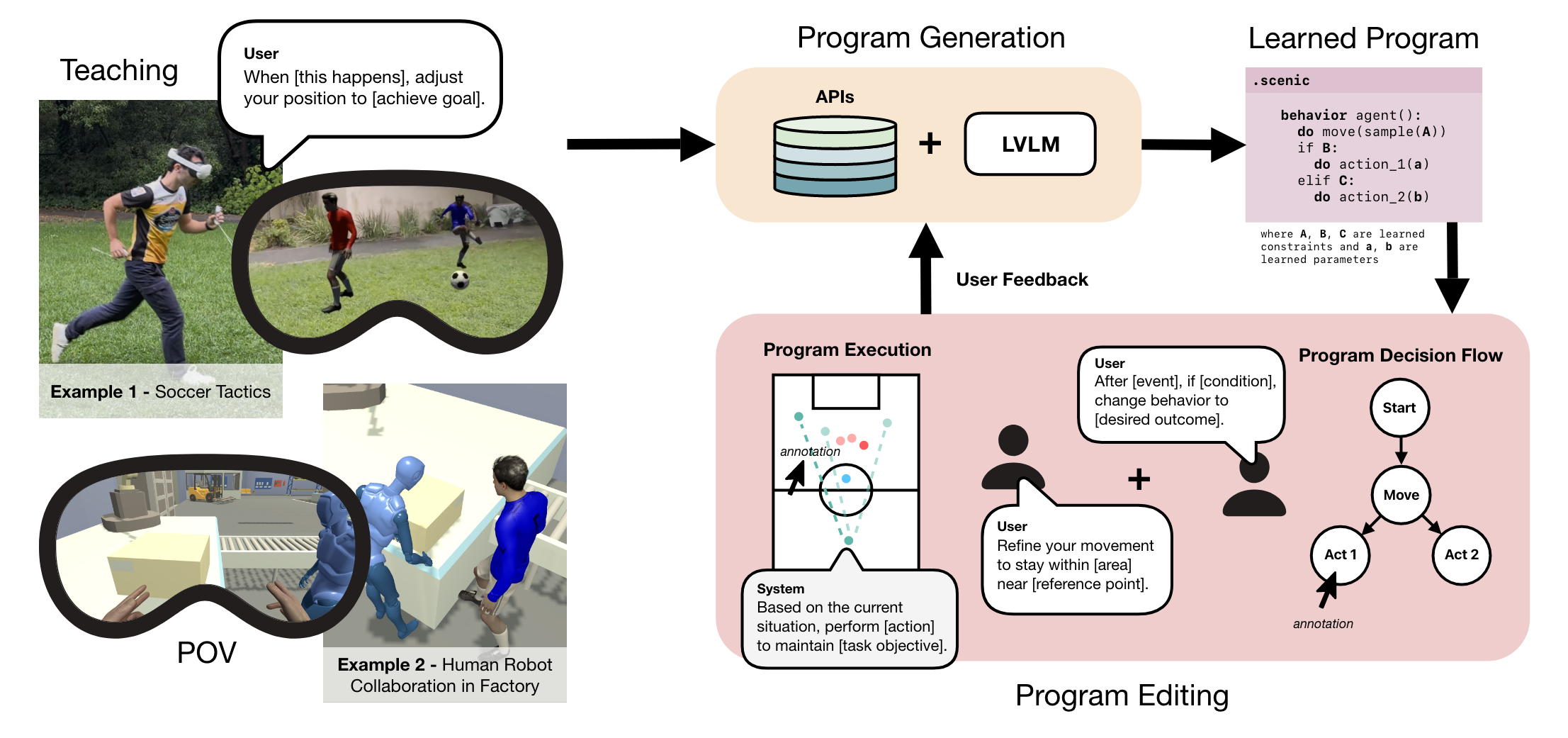}
  \Description{System architecture for interactively modeling collaborative physical activities as programs from narrated demonstrations.}
  \caption{Our system learns collaborative physical activities from narrated mixed reality demonstrations, generates programs with domain-specific APIs, and demonstrates back to the user for feedback and refinement.}
  \label{fig:teaser}
\end{teaserfigure}


\received{20 February 2007}
\received[revised]{12 March 2009}
\received[accepted]{5 June 2009}

\maketitle

\section{Introduction}
Enabling users to teach systems how to perform physical tasks has been a long-standing goal in human-computer interaction (HCI). While significant progress has been made in teaching individual skills, less attention has been given to enabling non-technical users to teach systems complex, collaborative activities, e.g., in sports, construction, and manufacturing. In this paper, we introduce a system that addresses several key challenges in this area. We aim to: (1) develop a system that can effectively acquire and model these collaborative activities in the physical world; and (2) enable users with no technical background to interactively teach and correct the system's learning to align its behavior with their intent. Addressing these problems would allow for the creation of systems that can scalably teach, assist, and guide people in complex, real-world collaborative scenarios in the future.

However, realizing such a system presents significant technical challenges. First, learning collaborative activities entails substantially greater complexity compared to single-user tasks.. In HCI, systems for learning physical tasks have predominantly relied on demonstrations and, more recently, natural language. While teaching via demonstrations can capture the nuances of physical movement, it often fails to infer the underlying logic or user intent~\cite{cypher1993watch}. Conversely, natural language can articulate high-level goals but struggle to ground abstract commands into specific, continuous trajectories in the physical world—a classic instance of the symbol grounding problem~\cite{harnad1990symbol,bisk2020experience}. Collaboration further increases this complexity since user's assumptions on collaborators' intent also need to be accounted for. Also, the constraint that the system must be teachable by users without a technical background necessitates that its internal models of physical activities be both interpretable and directly editable through an interface.

The existing literature on systems for representing collaborative physical activities is sparse. The most closely related works, such as GhostAR~\cite{cao2019ghostar} and VisCourt~\cite{Cheng2024VisCourt}, leverage mixed reality (MR) to generate virtual avatars for demonstrating interactions. This approach provides the system with a fully observable view of the behaviors of, and interactions between, the user and their collaborators. However, rather than teaching collaborative tasks, users \textit{configure} the interactions to generate tutorials. For example, in VisCourt~\cite{Cheng2024VisCourt}, a basketball coach selects and parameterizes pre-defined tactical plays from a menu-based interface. In our problem of interest, we aim to have users \textit{teach} collaborative activities to the system using natural modalities (e.g., language and demonstrations). 

We present an interactive system that enables non-programmers to model collaborative physical activities by synthesizing programs from {\em narrated demonstrations}. Our choice of this input modality—physical demonstrations paired with concurrent natural language—is informed by prior work showing that it facilitates more accurate and robust learning \cite{tung2018narrated,waytowich2019narration,yu2023deltaco}; narration disambiguates the intent behind an action, while the demonstration grounds language in the physical world. As summarized in Fig.~\ref{fig:teaser}, our system treats narrated demonstration as a key interaction primitive across the entire teaching loop—for initial instruction, inspection, and correction—enabling users to align the program's behavior with their intent without interacting with raw source code. The system allows a user to first teach a task via a narrated demonstration in mixed reality (MR). The system processes this recording with a large vision-language model (LVLM) to generate an initial program. The system communicates its learning via decision flow diagram (visualizing the underlying code structure) and program execution which provides narrated demonstrations to the user. To correct the program, the user provides feedback by providing narration along with physical demonstration via annotations. 

We evaluated our system in a within-subjects study with 20 non-programmer participants who used it to teach multi-player soccer tactics in a mixed reality environment. Our results show that users, leveraging the iterative feedback loop, were able to successfully teach and significantly refine the system's understanding of complex, collaborative behaviors to align with their intent. Users also reported ease of use in teaching and correcting the system. This paper makes the following key contributions:

\begin{enumerate}[label=(\arabic*)]
  \item An interactive programming system for modeling collaborative physical activities that uses narrated demonstrations as a unified modality for:
  \begin{enumerate}
      \item non-programmers to teach and correct the system, and
      \item the system to effectively communicate its learning to users.
  \end{enumerate}
  \item A user study in a tactical soccer setting, together with an existence proof in a factory environment, demonstrating the feasibility of our system for modeling collaborative physical activities as users intend.
\end{enumerate}

\section{Related Work}\label{sec:related_work}
\subsection{Tutorial Authoring Systems}
Augmented Reality (AR) has long been envisioned for delivering tutorials and task guidance. Seminal systems demonstrated knowledge-based overlays for step guidance \cite{Feiner1993KnowledgeAR} and proactive, sensor-driven instructions for physical assembly \cite{Antifakos2002Proactive}. However, these early systems focused on the \emph{delivery} of guidance, with authoring relegated to experts rather than end users. A significant thread in HCI research has focused on empowering practitioners to author their own tutorials. For example, InstruMentAR~\cite{Liu2023InstruMentAR} and AdapTutAR~\cite{Huang2021AdapTutAR} enable experts to record embodied demonstrations to produce step-wise AR overlays for machine operation, with AdapTutAR further tailoring guidance to the learner's state. Beyond AR/VR, the community has explored auto-authoring from demonstrations in mixed media. TutoriVR~\cite{Kumaravel2019TutoriVR} embeds 2D videos within VR workspaces accompanied by 3D contextual cues, while VAAR~\cite{Yamaguchi2020VAAR} reconstructs a 3D assembly graph from 2D video to align AR animations. TutorialLens~\cite{Kong2021TutorialLens} advances this by directly authoring AR tutorials from synchronized narration and demonstration, capturing finger trajectories and speech to structure the steps.

A critical distinction in recent work is the role of natural language. While much prior work simply records and replays narration to augment steps, newer systems analyze language to infer procedural structure. For instance, Truong et al. automatically derive two-level hierarchical procedures from instructional videos \cite{Truong2021Auto}, and TutoAI~\cite{Chen2024TutoAI} systematizes pipelines to extract tutorial components from videos and transcripts to assemble mixed-media guides. Complementing these authoring tools, recent work demonstrates fully automated AR guidance from Large Language and Vision Models (LLVMs), enriching steps with embedded visual cues \cite{Zhao2025GuidedReality}. While this highlights the increasing integration of modern AI, its focus is on automated generation rather than end-user \emph{teaching} and refinement.

Despite these advances, most systems target single-user, non-collaborative tasks. They seldom model role-dependent logic, joint timing, or the state of teammates and opponents. While systems like GhostAR \cite{cao2019ghostar} and VisCourt \cite{Cheng2024VisCourt} explore multi-agent choreography and tactic authoring in AR and MR, they primarily focus on configuring pre-defined interactions rather than learning an editable, closed-loop policy from narrated demonstrations.

\subsection{Interactive Task Learning}
In Interactive Task Learning (ITL), the goal is for a machine to learn task knowledge from a human teacher, often incrementally and through interactions that build an interpretable representation \cite{Laird2017ITL}. Prominent HCI examples include ONYX~\cite{Ruoff2023ONYX}, which enables users to teach natural-language interfaces by aligning speech and demonstrations with system capabilities, and VAL~\cite{Lawley2024VAL}, which uses LLMs in a neuro-symbolic pipeline to acquire editable, hierarchical procedures from conversation. These systems excel at software-centric or simulated tasks, emphasizing symbolically interpretable knowledge and conversational repair (e.g., clarifications, confirmations, undo). However, they typically do not ground learning in continuous physical trajectories or reason about multi-agent coordination and timing—challenges central to collaborative physical activities.

\subsection{Task Learning with Feedback in Reinforcement Learning.}
A substantial body of reinforcement learning (RL) research studies how agents can learn tasks from explicit human feedback rather than hand-crafted reward functions. Surveys of interactive RL and human-in-the-loop methods catalog a wide range of feedback modalities, including scalar evaluative rewards, critiques, demonstrations, and natural-language guidance~\cite{Najar2021HumanAdviceSurvey}. Early work such as the TAMER framework treats human input as real-valued evaluative feedback delivered while the agent acts; the agent learns a reward model directly from this feedback and optimizes its policy accordingly~\cite{KnoxStone2008TAMER}. Deep TAMER extends this idea to high-dimensional visual domains using deep networks, showing that sparse human reward signals can successfully shape Atari agents~\cite{Warnell2018DeepTAMER}. Preference-based RL instead elicits pairwise comparisons over trajectory segments, training a reward model that explains human preferences and then optimizing the policy with respect to that learned reward~\cite{Christiano2017HumanPreferences}. More recent interactive imitation-learning methods combine multiple forms of feedback—corrective demonstrations, scalar evaluations, and implicit positive feedback—and use uncertainty estimates to actively query humans when clarification is most useful~\cite{Celemin2023CorrectiveEvaluative}. Together, this literature demonstrates that RL agents can leverage diverse feedback channels (evaluative rewards, preferences, and corrections) to learn complex behaviors with relatively few interactions, but typically focuses on single-agent tasks with relatively sparse literature in the space of collaborative physical activities.

\subsection{Programming by Demonstration with Corrective Feedback.}
Robot Programming by Demonstration (PbD), or Learning from Demonstration (LfD), aims to let users specify robot behaviors by showing examples rather than writing code~\cite{Billard2008PbDHandbook,Argall2009LfDSurvey}. Surveys and tutorials describe a spectrum of approaches that encode demonstrations as trajectories, movement primitives, finite-state machines, or probabilistic models, and highlight design choices in how demonstrations are collected (kinesthetic teaching, teleoperation, motion capture) and generalized to new situations~\cite{ChernovaThomaz2014RLFromTeachers,Calinon2019LfDEncyclopedia}. While many PbD systems treat demonstrations as fixed training data, some explicitly incorporate user feedback to refine learned programs. For instance, Argall et al.\ allow users to provide corrective feedback on low-level motion primitives, using this input to scaffold and refine policies built from previously demonstrated skills~\cite{Argall2011TeacherFeedback}. Such work shows that combining demonstration with post-hoc corrections can improve performance and data efficiency, but typically offers only limited support for inspecting or editing the learned task representation itself (e.g., at the level of high-level decision logic). Our work builds on these ideas by representing tasks as interpretable programs and designing interfaces that treat feedback, inspection, and correction as interaction mechanisms for modeling collaborative physical activities.

\subsection{Interactive Program Synthesis}
A parallel research thread focuses on synthesizing executable programs from end-user input, such as examples or demonstrations. Canonical work like FlashFill~\cite{Gulwani2011FlashFill} demonstrated how carefully designed domain-specific languages (DSLs) could yield powerful transformations from a few examples. Subsequent HCI research has developed interaction models to make synthesized programs more understandable and \emph{editable}. This includes techniques for disambiguating synthesized programs via navigation and active-example queries~\cite{Mayer2015DisambPBE}, and augmented-example workflows that allow users to iteratively repair programs~\cite{Zhang2020AugEx}. Prior work also shows that explicitly surfacing symbolic representations of program structure helps non-programmers inspect logic and make targeted edits, for example by exposing block-level units of generated code~\cite{yen2024coladder,jiang2022genline}, visualizing and monitoring execution traces of LLM-generated code~\cite{xie2024waitgpt}, or decomposing analysis tasks into editable subgoals that users can steer and verify~\cite{kazemitabaar2024interactive}.

Domain-specific systems show how end users can create and refine automations involving complex logic like loops and conditionals. Examples include Rousillon~\cite{Chasins2018Rousillon} for hierarchical web scraping, DiLogics~\cite{Pu2023DiLogics} for handling diverse input conditions, and MIWA~\cite{Chen2023MIWA} for mixed-initiative diagnosis and repair. In mobile computing, SUGILITE~\cite{Li2017SUGILITE} and PUMICE~\cite{Li2019PUMICE} combine natural language with programming-by-demonstration (PbD)~\cite{cypher1993watch} to synthesize procedures that users can inspect and modify. In data science, Wrex~\cite{Drosos2020Wrex} produces readable notebook code from examples, Falx~\cite{Wang2021Falx} synthesizes visualization pipelines from example outputs, and recent interfaces like SQLucid~\cite{Tian2024SQLucid} generate step-by-step, editable natural-language explanations to help non-experts understand and correct model-generated SQL.

\section{Background}
This work builds on recent advances in multimodal learning and probabilistic programming. In this section, we introduce the technical foundations that enable our system design.
\subsection{Large Vision Language Models}
Large Vision Language Models (LVLMs) extend large language models by processing visual inputs in addition to linguistic ones. LVLMs have demonstrated strong capabilities in aligning natural language descriptions with structured visual data, supporting tasks such as captioning, multimodal retrieval, and program synthesis \cite{radford2021clip,yuan2021florence,li2022glip}. Their capacity to integrate symbolic and perceptual representations makes them well suited for interpreting narrated demonstrations, which combine visual demonstrations of physical activity with verbal explanations. In our work, LVLMs serve as the program synthesis tool.

\subsection{Scenic Probabilistic Programming Language}
Scenic~\cite{fremont2019scenic,scenic-mlj23} is a domain-specific probabilistic programming language for modeling and generating physical environments with stochastic multi-agent behaviors and initial conditions in simulation. A snippet of a Scenic program is shown in Fig.~\ref{fig:program}. When compiling such a program, Scenic imports (i) the map of the simulated world, (ii) definitions of objects (e.g., \texttt{ball}, \texttt{goal}) and agents (e.g., \texttt{Coach}, \texttt{Teammate}), and (iii) an API library defining the action space (e.g., \texttt{MoveTo}, \texttt{Pass}, \texttt{Speak}) and support functions (e.g., \texttt{HasBallPossession()}). Using this information, the environment is modeled as follows. Lines 22–28 specify a distribution over initial conditions by instantiating the user and other agents with initial positions and assigned behaviors. Because the world map has been imported, coordinates are interpreted relative to the map origin. The user’s behavior (\texttt{CoachBehavior()}) is modeled using the provided APIs. At execution time, Scenic samples an initial condition for each agent: the user’s location is fixed at the map origin, while the teammate’s and opponent’s positions are sampled from specified ranges relative to the user’s position and orientation. After instantiating objects and agents, Scenic executes each agent’s behavior function to generate a dynamic, reactive environment.

\section{System Architecture}\label{sec:system_architecture}
An overview of our system architecture is illustrated in Fig.~\ref{fig:teaser}. Throughout this section, we will use a running example where user teaches a soccer tactic on how to lure (or attract) an opponent defender away from the teammate to create space for the teammate to score.

\subsection{Teaching Interface in Mixed Reality}\label{sec:teaching-ar-interface}
\textbf{Environment Modeling.} To model and generate MR environments involving multiple entities with interactive behaviors, we use Scenic~\cite{fremont2019scenic}, a probabilistic programming language designed to specify distributions of physical scenarios. \\
\textbf{Action Space.} We assume that the action space is provided. These actions (e.g., dribbling, kicking) can be executed by either users physically moving or using MR controllers. For example, the user may physically move around to re-position oneself, but for kicking a virtual ball, one needs to use the controller to execute this action.\\
\textbf{Embodied Narrated Demonstrations.} The user wears a mixed reality (MR) headset and is immersed in a 3D workspace, embodying the specific agent they aim to teach. While embodying this agent, the user authors the initial scene and assigns behaviors to surrounding agents to incrementally build a dynamic environment. In the running soccer example, the user first constructs the scene by placing a teammate, an opponent, goalposts, and a ball at desired locations via verbal instructions and MR controller input (e.g., ``Place my teammate here'' [user points at (12.3, 4.26)]). The user then verbally assigns behaviors to these surrounding agents (e.g., ``The teammate should pass the ball to me.''). A large language model is used in real time to map verbal instructions and controller annotations to actions from a provided action space, identify which agent or object each action targets, and immediately execute the corresponding actions to generate the instructed environment. As the user incrementally builds this dynamic environment, one provides narrated demonstrations of how the focal agent should react to and coordinate with surrounding agents. Unlike surrounding agents, the user can enact the focal agent’s behavior through physical demonstrations while simultaneously narrating and annotating via the controllers. During this teaching phase, the MR headset records: (1) RGB video of the user and surrounding agents and objects from a virtual camera, (2) timestamped annotations of actions triggered with MR controllers (e.g., [12:33] User passed to Teammate''), and (3) a transcription of the narration as timestamped tokens (e.g., [12:34] Therefore’’, [12:35] ``you’’).
\begin{figure}[H]
\centering
\includegraphics[width=1.0\linewidth]{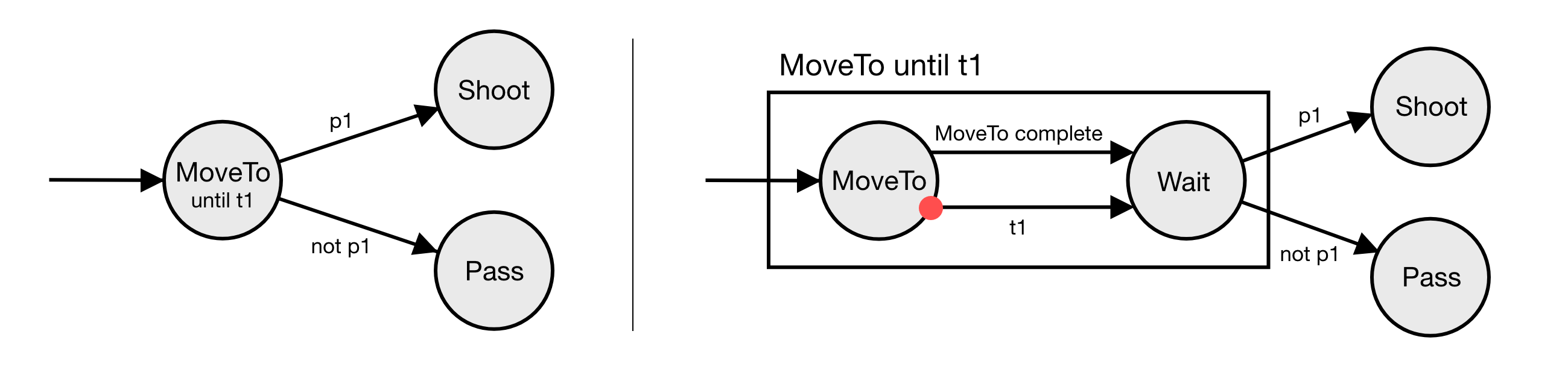}
\Description{Finite state machine diagram showing actions, waiting states, and transitions based on preconditions and termination conditions.}
\caption{Finite state machine representation of behaviors for our running example. The left figure shows a hierarchical FSM modeling a set of narrated demonstrations, and the right figure zooms into an hierarchical state of the FSM.}
\label{fig:fsm}
\end{figure}

\subsection{Program Modeling}\label{sec:program-modeling} 
\subsubsection{Modeling Behavior as a Finite State Machine}\label{sec:fsm}
The objective of program generation in our system is to model the user's behavior from narrated demonstrations. We formally represent the behavior as a hierarchical, interrupt-driven finite state machine (FSM), a standard abstraction in program generation that compactly encodes contingencies via conditional branching. In our setting, states (nodes) denote actions, while directed edges denote preconditions—Boolean predicates over continuous variables (e.g. positions, orientations) of the physical world—that gate transitions. This abstraction allows the model to capture both discrete sequencing and coordinated behavior conditioned on the physical world state. 

Returning to our running example, the FSM in Fig.~\ref{fig:fsm}(left) models two narrated demonstrations from the user teaching how to lure an opponent. The user's narrates ``So, in this scenario you need to lure the opponent to either sides of the field like this to receive a pass from your teammate. After you receive the ball and the opponent does come after you, then your teammate is open so pass to your teammate.'' In the next narrated demonstration, user states ``After you receive the ball, if the opponent does not budge and goes after your teammate, then you can shoot for the goal.'' The FSM in Fig.~\ref{fig:fsm}(left) models the decision flow of the user in \textit{both} of the narrated demonstrations as a FSM.

\textbf{Hierarchical vs. Primitive States.} There are two types of states: hierarchical and primitive. An hierarchical state represents a FSM which consists of primitive states that does not expand into its own FSM. In Fig.~\ref{fig:fsm}(left), we represent a behavior with hierarchical states whose transitions are modeled with the preconditions over the edges. Each hierarchical state represents a FSM consisting of primitive states and edges, which represents a micro-controller that executes the action represented by the hierarchical state. Fig.~\ref{fig:fsm}(right) zooms into \texttt{MoveTo until t1} state in Fig.~\ref{fig:fsm}(left), where \texttt{MoveTo} is the primitive state that executes the action followed by an interrupt-driven and non-interrupt-driven edges. If the action \texttt{MoveTo} completes, then it transitions to the \texttt{Wait} state which idles and takes no other action until either preconditions \texttt{p1} or \texttt{p2} are satisfied. If a termination condition (\texttt{t1}) is satisfied while \texttt{MoveTo} action is in progress, then an interrupt occurs where the transition to \texttt{Wait} occurs regardless of whether \texttt{MoveTo} completes. This interrupt-driven transition models the contingent human behavior that reacts to dynamically unfolding physical environment. The FSM in Fig.~\ref{fig:fsm} can be modeled as a Scenic program in Fig.~\ref{fig:program}.

\subsubsection{Library of APIs}\label{sec:apis}
The library of APIs defines the action space and models physical constraints, both of which are domain-specific and need to be tailored accordingly. 

\textit{Constraint APIs for Logical and Spatial Reasoning.} To logically and spatially reason about the state of the physical world, our library provides a set of \emph{constraints}. We define a constraint as an API that returns either a boolean value or a spatial field over the workspace, indicating whether the current state of the world satisfies the constraint, or where in the workspace the constraint would hold. 

For instance, a constraint \texttt{DistanceTo(obj, ref, d, operator)} returns \texttt{True} if object \texttt{obj} is at distance \texttt{d} from \texttt{ref} under the specified operator, or alternatively a probability distribution over the workspace indicating where \texttt{obj}'s location would satisfy the constraint. This constraint is useful in our running example for detecting when the opponent is applying pressure to the user—for instance, when the opponent is positioned too close.


\textit{Composition of Constraint APIs.} Constraints, whether boolean or spatial, can also be composed with logical operators: AND, OR, and NOT. Boolean composition follows standard logic, returning a final \texttt{True} or \texttt{False}. For spatial fields, composition is defined as:
\[
A \land B = A \cdot B, \quad
A \lor B = A + B, \quad
\lnot A = 1 - A,
\]

where $A$ and $B$ are constraint spatial fields. The resulting distribution is a probability distribution discretized over the workspace, following the logical construction of the constraints. After normalization, this distribution can be sampled to produce concrete positions. 

\begin{figure}[H]
\centering
\includegraphics[width=0.9\linewidth]{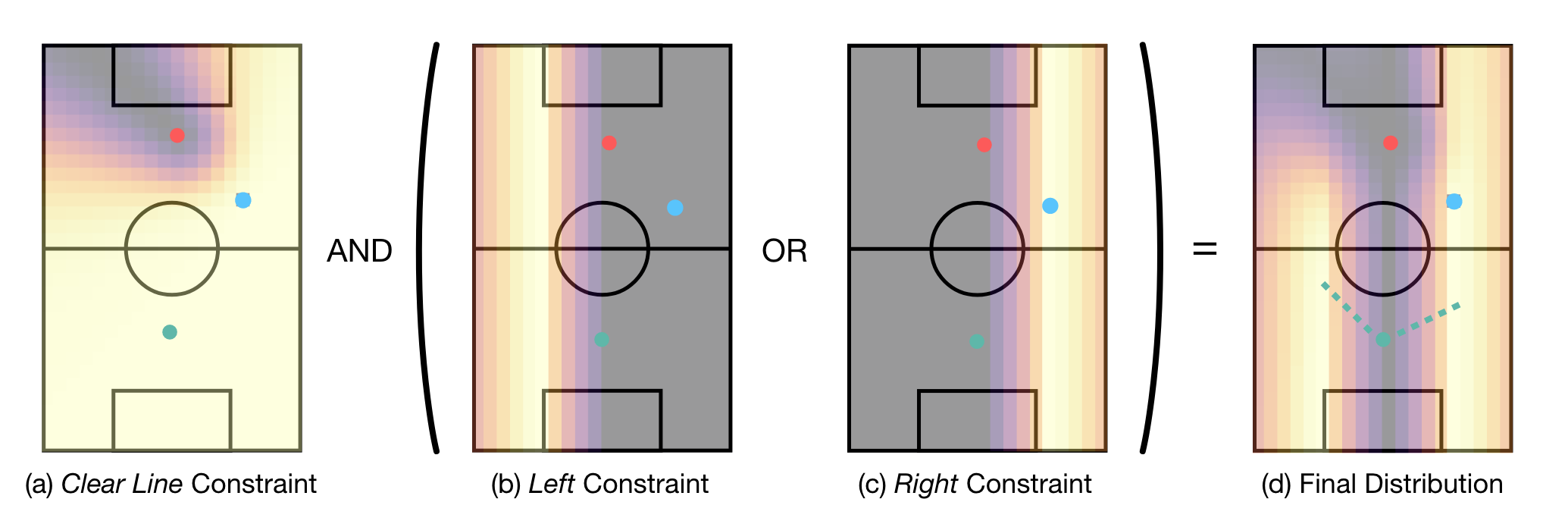}
\caption{Composing spatial constraints in the running example. Each panel shows a probability distribution over the field: (a) availability of a clear passing line, (b) positioning to the left of the defender, and (c) positioning to the right. These constraints are composed with logical operators (AND/OR) to produce the final distribution (d), which is sampled to determine where the agent should move to create space for a teammate’s pass as taught by the user.}
\Description{A figure illustrating how spatial constraints, visualized as probability distributions, are logically combined to yield a final distribution that guides agent positioning.}
\label{fig:constraints}
\end{figure}

Returning to our running example, the user taught ``So, in this scenario you need to lure the opponent to either sides of the field like this to receive a pass from your teammate.'' Fig.~\ref{fig:constraints} illustrates how spatial constraints can be logically composed to capture the conditions in our running example. To receive a pass from the teammate (blue), the user should not be standing behind the opponent (red). Thus, Fig.~\ref{fig:constraints}(a) assigns low probability (dark area) distribution to locations behind the opponent from the teammate's perspective, while assigning high probability on all other locations (yellow). To move to either sides, the two constraints (Fig.~\ref{fig:constraints}(b) and (c)) model horizontal conditions assigning high probability distributions to either sides of the field. The gradation of colors reflect the Gaussian distribution which we encoded to these constraints. The composition of these constraints means applying the athematic shown above followed by a normalization as visualized in Fig.~\ref{fig:constraints}(d).

\paragraph{Action APIs}

Action APIs define the action space of a user which can be either discrete (e.g. Pass) or continuous (e.g. MoveTo(destination)).

\subsection{Program Generation}\label{sec:program_synthesis}

\subsubsection{Data Processing}
\label{sec:data-preparation}
We process three types of data collected during the teaching phase: (1) RGB video captured from a virtual camera, (2) timestamped action annotations triggered via MR controllers, and (3) narrated descriptions transcribed as timestamped tokens. The video is treated as a sequence of timestamped image frames. Inputting all frames directly to the LVLM could exceed its input token budget. In such a case, we sub-sample the sequence (e.g., selecting a frame every second) to preserve the temporal structure of the demonstration while keeping the number of images within the LVLM’s context limits.

We then ground the timestamped actions into the transcribed text so that the resulting transcript encodes both narration and user actions in a temporally synchronized form. For instance, in our running example, the grounded transcript becomes: “You should move to either side of the field [user marked coordinate (12.3, 2.45)] to open up for a pass. [Teammate1 passed to user] Then shoot if a clear line is available [user shoots toward the goal]; otherwise, pass back to your teammate.” Grounding is achieved by aligning timestamps of word tokens with logged events.

\begin{figure}
    \centering
    \includegraphics[width=.75\linewidth]{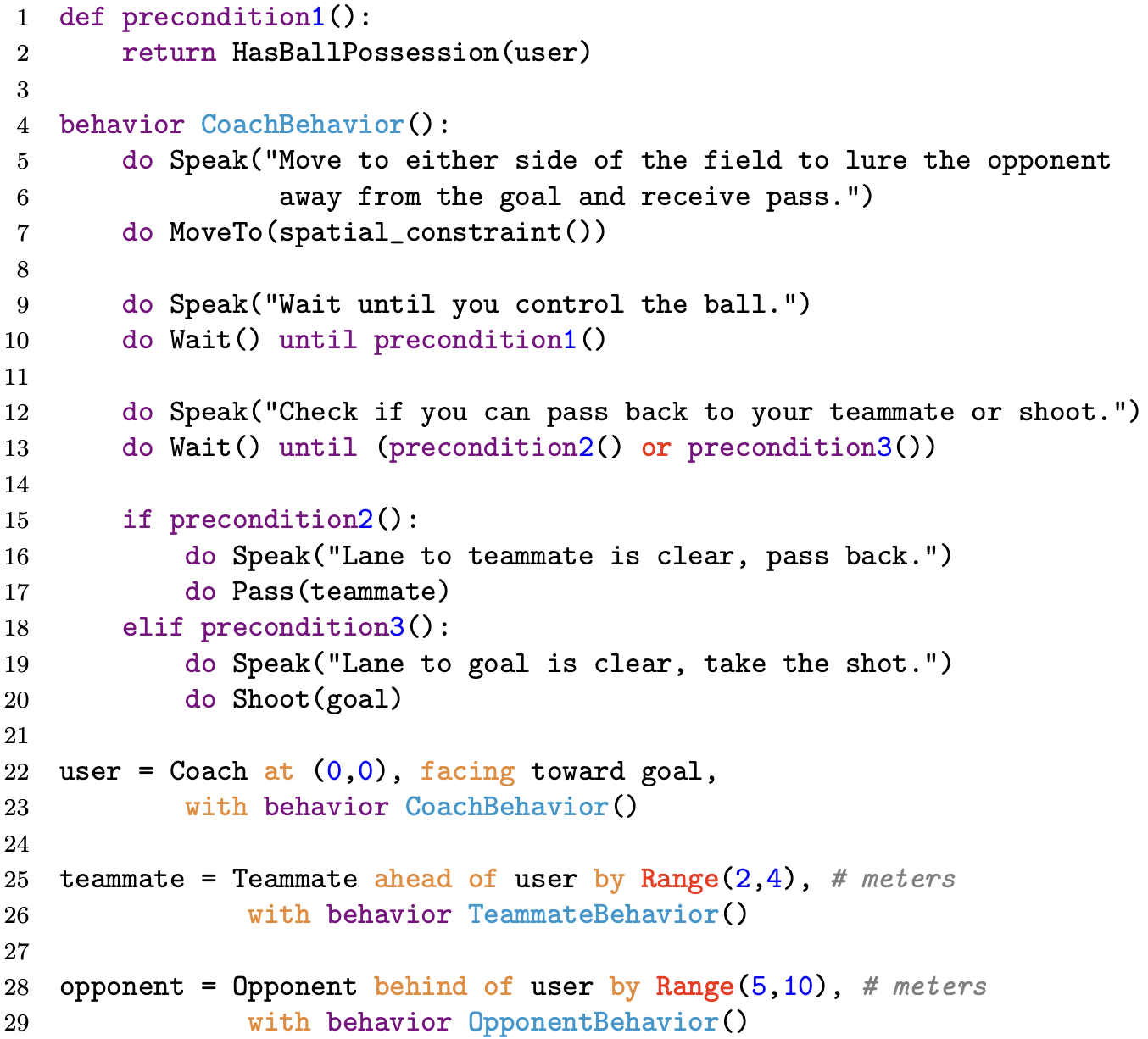}
    \caption{A snippet of a Scenic program modeling the user’s behavior when interacting with a teammate and an opponent in the running soccer example. The actions, constraints, and preconditions are expressed using a provided API library that defines the action space and physical constraints described in Sec.~\ref{sec:apis}.}
    \label{fig:program}
    \Description{Example Scenic program for the coach behavior and constraints in the lure scenario generated by our system.}
\end{figure}

\subsubsection{Program Generation}\label{sec:program_generation}
The LVLM takes the following as input: (1) pairs of sub-sampled images and their corresponding grounded transcripts, (2) textual annotations of the available APIs, (3) documentation of the syntax and semantics of the target programming language, and (4) a system prompt. The system prompt provides a few example programs and instructs the LVLM to generate a Scenic program that models the user’s instructions using the provided APIs. For concrete details of these inputs, please refer to Supplement~\ref{appendix:system_implementation}. Through this query, the LVLM generates a full executable Scenic program, instantiating agents and objects and assigning initial conditions and behaviors to agents.

\textit{Adding Temporally Aligned Narration to System Demonstration.}
Demonstrations alone can be ambiguous, as users may infer multiple, conflicting interpretations of the system’s intent. To address this, we introduce a \texttt{Speak()} API that allows the program to narrate its intent during execution. This narration occurs in real-time, synchronized with the corresponding actions, enhancing interpretability. Prior work in interactive program generation has explored natural language explanations through inline annotations~\cite{angert2023spellburst} or step-by-step summaries~\cite{liu2023whatitwantstosay}. Building on this, our system prompts the LVLM during generation to insert line-by-line rationales using \texttt{Speak(text)} calls, which encode the reasoning behind each local action. At runtime, each \texttt{Speak} statement briefly pauses the simulation, plays the narration via a text-to-speech model, and then resumes execution. Because these narrations are embedded directly within the program’s structure, they are temporally aligned with the behavior being demonstrated, enabling users to understand not just what the system is doing, but why it is doing it, in the moment. Returning to our running example, the generated program modeling the user behavior with narration is shown in Fig.~\ref{fig:program}(left). 


\begin{figure}
\centering
\includegraphics[width=1.0\linewidth]{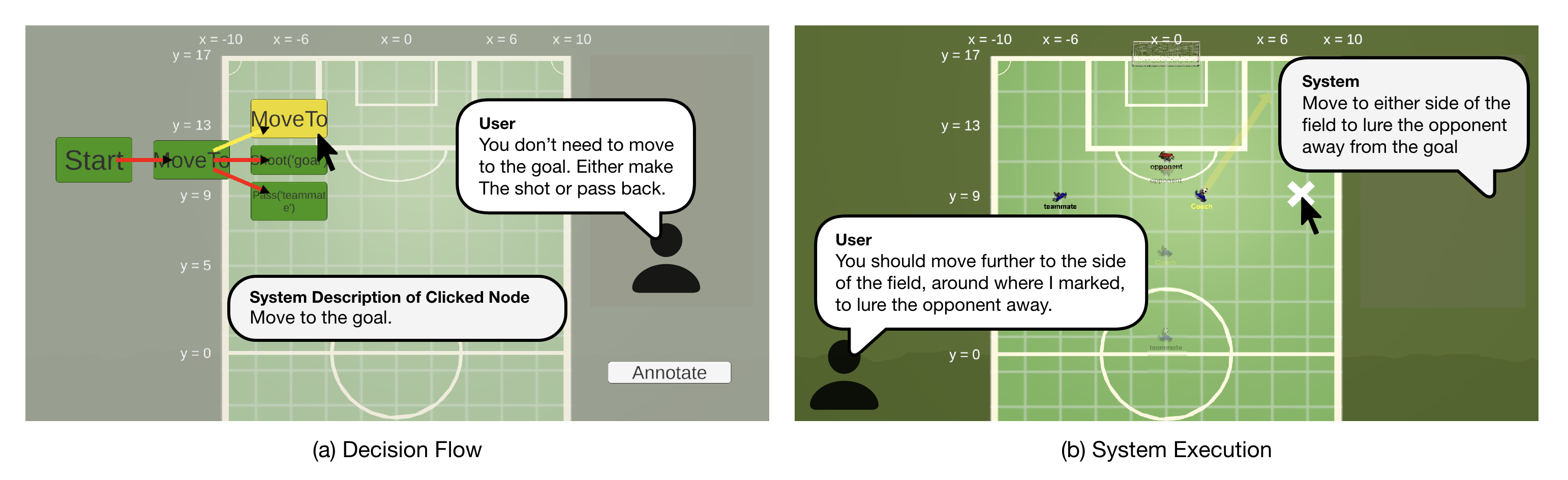}
\Description{Illustration of user-in-the-loop feedback with Decision Flow and System Execution modes.}
\caption{User-in-the-loop feedback is supported through two complementary modes: 
(a) \textbf{Decision Flow}, which presents a finite state machine derived from the generated code, enabling structured inspection and annotation of program logic. Users are able to click and highlight edges and nodes by using "Annotate." When edges and nodes are highlighted yellow through annotation, our system will know which edgers and nodes need to be fixed. 
(b) \textbf{System Execution}, which replays the modeled behavior under varied conditions with optional narrated \texttt{Speak()} actions, allowing users to validate fidelity in practice and provide situated corrections. Users can provide annotations in the environment by clicking on the field, leaving behind an X mark.
Together, these modes balance interpretability and experiential validation, ensuring programs remain both logically consistent and behaviorally faithful.}
\Description{Illustration of user-in-the-loop feedback with Decision Flow and System Execution modes.}
\label{fig:feedback}
\end{figure}

\subsection{User Inspection and Feedback to System}\label{sec:correction}

Our method allows users to amend the program through two primary code-free editing interfaces.

\subsubsection{Decision Flow}
\label{sec:decision-flow}

Prior work shows that symbolic representations of program structure help non-programmers inspect logic and make targeted edits~\cite{xie2024waitgpt, yen2024coladder, jiang2022genline, kazemitabaar2024interactive}. We adopt this visualization approach by displaying the program’s FSM. For the program in Fig.~\ref{fig:program}, the user is shown the corresponding decision flow chart in Fig.~\ref{fig:feedback}(a). Nodes and edges include descriptions that users can reveal by clicking, which convey the rationale and underlying code logic without exposing the code itself. Users can annotate selected nodes and edges while providing verbal feedback, enabling direct edits to the program’s structure. The user might look at the decision flow chart in Fig.~\ref{fig:feedback}(a) and annotate the node for \texttt{MoveTo} and explain that "You don't need to move to the goal. Either make the shot or pass back." Otherwise, if everything was fine with the decision flow chart, the user could just verbally state that "no modifications are needed."

\subsubsection{Program Execution}
\label{sec:execution}

The program runs as a Unity MR simulation, producing an embodied, narrated execution of the learned behavior. It controls an avatar—a proxy for the user—across variations of the environment encountered during teaching. As the behavior unfolds, the rationale and actions are narrated through the \texttt{Speak} actions defined in the code as seen in Fig. \ref{fig:program} (left). Users can pause at any time, point and click to annotate the environment, provide verbal feedback, and then resume—supporting multiple corrections within a single run. As in data processing for program generation, we capture both video and a grounded transcript of the user’s feedback, with interleaved annotations. Incorporating the \texttt{Speak()} outputs into the transcript further helps the LVLM localize feedback to specific points in the program logic, using the narration as a debug print statements. After looking at the program execution in Fig. \ref{fig:feedback}(b) the user might pause at the frame shown and verbally explain that "You should move further to the side of the field, around where I marked, to lure the opponent away [user annotated \((x, y)\)]."


\subsubsection{Program Repair}
We then provide the multimodal stream of feedback data including videos, annotations and grounded transcripts to the LVLM together with the candidate program to amend and the user's initial narrated demonstration. The original prompt for program generation is augmented to instruct the LVLM on how to utilize feedback data and make localized edits. 


\section{Generalizability of Our Approach}
\begin{figure}
    \centering
    \includegraphics[width=0.7\linewidth]{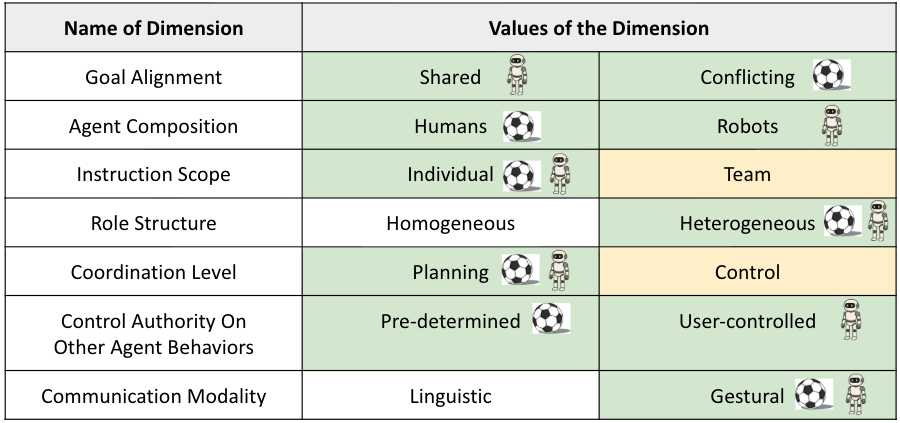}
    \caption{Design space of collaborative physical activities along seven dimensions. Soccer-ball and robot emojis mark values instantiated in soccer and factory examples ( Fig.~\ref{fig:feedback} and \ref{fig:factory}), respectively; yellow cells indicate dimension values not supported by our current system, while green cells indicate dimension values that are supported.}
    \label{fig:generalizability}
\end{figure}

\textbf{Overall scope of activities.}
Our approach can model a broad class of collaborative physical activities in which an individual agent’s behavior is represented as a high-level policy over physical states and actions. It reasons about how a focal agent should react to or coordinate with others by planning and sequencing such actions (e.g., “dribble,” “grasp an object,” “move into open space”). However, it \emph{cannot} model the underlying low-level motor skills that implement these actions, which require fine-grained coordination among joints with many degrees of freedom. It is unclear how to expose this kind of motor-domain knowledge as reusable APIs; for example, specifying how finger joints should coordinate in 3D to grasp objects with diverse shapes, sizes, and tactile properties. Without APIs that encode this level of detail, it is difficult to learn such coordination from only a few narrated demonstrations, and learning such fine-grained motor skills from even large datasets remains an open challenge in robotics. Furthermore, our approach is not designed to model activities that aim to teach a team of collaborators as a single unit. For user inspection (Sec.~\ref{sec:program_generation}), the system generates narrated demonstrations that explain the rationale behind the actions of a \emph{single} agent; it is not designed to model or explain collective coordination strategies across multiple agents. In practice, this means our approach is applicable for modeling high-level policies of \emph{individual} agents acting within collaborative activities, given a predefined low-level action space, but not for learning the underlying motor skills or synthesizing team-level control policies.

\textbf{Dimensions of collaborative activities.}
To more precisely characterize the class of activities our system targets within this scope, we adopt seven dimensions of collaborative activities grounded in prior work (Table~\ref{fig:generalizability}). \emph{Goal Alignment}~\cite{Deutsch1949,JohnsonJohnson2005} captures whether the task goal is shared among collaborators (e.g., manufacturing) or conflicting (e.g., one-on-one boxing). \emph{Agent Composition}~\cite{DeSantis2008,OnnaschRoesler2021} distinguishes collaboration among humans or robots. \emph{Instruction Scope}~\cite{MaloneCrowston1994,Gervasi2020} specifies whether the task targets an individual collaborating with others or a team working collectively. \emph{Role Structure}~\cite{Belbin1981,Belbin1993} characterizes whether collaborators’ roles are identical (e.g., synchronized dancing) or heterogeneous (e.g., soccer). \emph{Coordination Level}~\cite{DeSantis2008,MaloneCrowston1994} indicates whether the collaboration requires coordination at the planning level or the control level. Here, control refers to low-level actions that require coordinating multiple body joints with high degrees of freedom (e.g., grasping or throwing a ball), while planning sequences these low-level actions to achieve a task. \emph{Control Authority}~\cite{Parasuraman2000,Sheridan1992} refers to who controls collaborators’ behaviors, ranging from pre-determined policies (e.g., pre-determined strategies of soccer opponents) to real-time user control. Finally, \emph{Communication Modality}~\cite{OnnaschRoesler2021,Gervasi2020} specifies whether interaction relies on linguistic or gestural communication among agents (e.g., raising a hand to signal another agent). Combinations of values within each dimension are also possible but are omitted from the table for brevity.

\begin{figure}
    \centering
    \includegraphics[width=\linewidth]{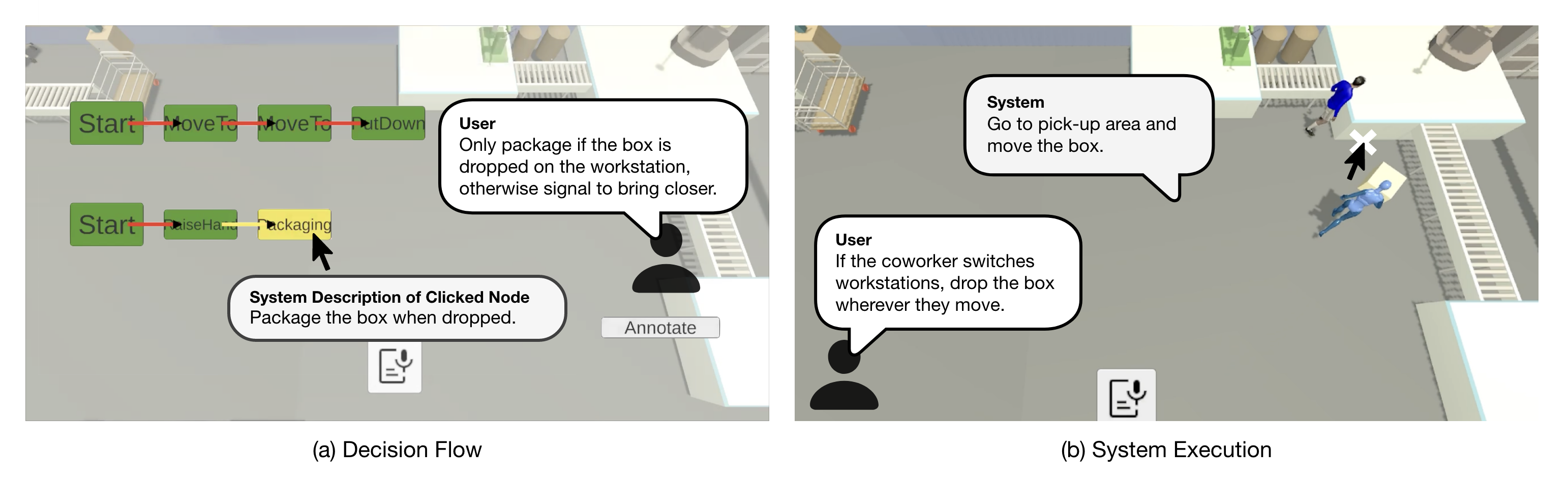}
    \caption{User feedback on decision flow and system execution in a factory manufacturing setting where a human and a robot collaborates to package a box.}
    \label{fig:factory}
\end{figure}

\textbf{Instantiated examples: soccer and factory.}
Within this design space, we instantiated our system in two concrete domains. The running soccer example in Sec.~\ref{sec:system_architecture} models multi-agent tactics in which a focal player coordinates with teammates and opponents at the level of high-level decisions (e.g., luring defenders, creating passing lanes). To further demonstrate generalizability, as illustrated in Fig.~\ref{fig:factory}, we instantiated our system architecture in a factory manufacturing setting in which a mobile robot collaborates with a human worker to package a box. In this scenario, the user authors the interaction by instantiating the environment (placing the human, robot, and box) and specifying initial conditions and behaviors for each agent. For example, the user can define that the human begins by raising a hand to gesture to the robot, and that the robot should interpret this gesture as a command to fetch a box, deliver it to the human, and enable the human to complete the packaging task. Following the procedures described in Sec.~\ref{sec:system_architecture}, the user then teaches, inspects, and corrects the programs modeling this collaboration. We present an existence-proof with three participants, all of whom were able to author and iteratively refine collaborative activities, yielding programs that accurately modeled their desired human--robot interactions. The results are included in Supplement~\ref{supplement:factory}.

In Table~\ref{fig:generalizability}, the cells highlighted in green and marked with soccer-ball or robot emojis indicate which dimension values are instantiated in the soccer and factory examples, respectively. Cells where both values of a dimension are highlighted in green indicate that our approach can model combinations of values, which are omitted in the table.

\textbf{Unsupported and untested regions of the space.}
Cells highlighted in yellow denote dimension values that our system cannot currently model. Cells that are not colored correspond to dimension values that we did not empirically evaluate in this paper but for which we expect our approach to generalize. For example, homogeneous role structures such as synchronized dancing could be modeled by assigning the same learned behavior to all agents in the activity. Likewise, linguistic communication among agents could be supported by using large language models to parse utterances and trigger appropriate coordinated actions.

\section{User Study}

We aim to evaluate the efficacy of our system in supporting users with no programming background to accurately externalize their domain knowledge of collaborative physical activities as programs. In particular, we investigate the 
four research questions in Table~\ref{tab:research-questions}.

\subsection{Participants}\label{sec:participants} The protocol for our study was approved by our institution's IRB. We recruited 20 participants via email to our institution’s intramural, club, and varsity soccer groups. Inclusion criteria required competitive soccer experience (at least competitive intramural, club, or varsity at the high-school or college level). We excluded individuals who only played recreational leagues, had a history of seizures, visual abnormalities, susceptibility to motion sickness or nausea, scalp skin irritation, or used wearable medical devices, consistent with the MR headset's safety guidance. The signed informed consent form was obtained for each subject via email or in person prior to the study. Each participant received a \$30 gift card.

\begin{table}[th]
\centering
\caption{Research questions and measures.}
\label{tab:research-questions}
\renewcommand{\arraystretch}{1.2}
\begin{tabularx}{\textwidth}{@{}>{\raggedright\arraybackslash}X >{\raggedright\arraybackslash}p{0.34\textwidth}@{}}
\toprule
\textbf{Research Questions} & \textbf{Metrics} \\
\midrule
\addlinespace

\textbf{RQ1:} Do users find it intuitive to teach physical activities via narrated demonstrations with our embodied MR interface? &
\begin{tabular}[t]{@{}l@{}}1.\ User Questionnaire\end{tabular} \\
\hline

\addlinespace
\textbf{RQ2:} Can users generate, on first attempt, an accurate program that models their behaviors? 
&
\begin{tabular}[t]{@{}l@{}}1.\ Correctness score\\ 2.\ Completeness score\end{tabular} \\
\hline

\addlinespace
\textbf{RQ3:} Can users efficiently improve the accuracy of the program via iteratively editing with our system? &
\begin{tabular}[t]{@{}l@{}}1.\ Correctness score\\ 2.\ Completeness score\end{tabular} \\
\hline

\addlinespace
\textbf{RQ4:} Do users find it intuitive to edit the program? &
\begin{tabular}[t]{@{}l@{}}1.\ User Questionnaire\end{tabular} \\
\bottomrule
\end{tabularx}
\end{table}

\subsection{Experiment Setup}
Users used a Meta Quest 3 MR headset to teach soccer tactics at outdoor on a lawn (Fig.~\ref{fig:teaser}). When teaching soccer tactics in MR, users moved around and executed actions (e.g. pass, shoot, start/stop recording, pause/unpause scenario) using MR controllers. After users provided narrated demonstrations, the headset transmitted the collected data to the study personnel's laptop. All program synthesis were conducted on the laptop. OpenAI 5.0-mini was used as LVLM. The decision flow diagrams and the program executions were also shown in the laptop to collect user feedback. 


\subsection{Study Design}
We conducted a within subjects study where participants evaluated the system's learning before and after they provided corrective feedback. We chose soccer as an example domain for complex collaborative physical activities as it involves highly dynamic coordination with multiple players. Three different soccer scenarios (luring opponent, overlapping teammate, distributing pass) were prepared by the study personnels (refer to Supplement~\ref{supp:user_scenarios}), where the behaviors of the environment agents (e.g. teammates, opponents) were pre-scripted. These three scenarios are derived and adapted from the England Football Learning~\cite{englandfootball-sessions}, a popular website for youth soccer coaches regarding soccer drills. Each participant embodied the coach avatar in MR to provide narrated demonstrations. The soccer scenarios were assigned to participants in a round-robin manner.

\subsection{Study Procedure}\label{sec:study_procedure}
Each participant completed a 60-minute one-on-one session. After watching a tutorial video (covering how to teach in AR, inspect, and correct system learning), participants had a hands-on tutorial in MR to familiarize the buttons on the controllers. Then, they watched a video informing them of their assigned scenario and the pre-scripted behaviors of the environment agents (e.g. teammates, opponents), and reviewed the rubric that would be used for evaluation of the learned programs later to ensure that their teaching covers the rubric. This rubric per scenario specified objective task requirements, such as scoring a goal and passing to a teammate (refer to Supplement~\ref{supp:rubric}). They then recorded two narrated demonstrations of the same tactical goal (re-recording allowed). The system synthesized a program and displayed a decision-flow diagram; if issues were identified, participants were allowed to provide at most one corrective-feedback and the LLM repaired the program. Next, the system showed three narrated demos of its learned behavior; if problems remained, a fourth demo was shown and participants were allowed to provide at most one feedback, prompting another repair. The restrictions on the number of user feedback was enforced due to limited study time. Furthermore, we asked users to provide feedback that was consistent with their initial narrated demonstrations, even if they came up with better tactics over time, so that we could clearly measure improvement before and after feedback within the limited session. Participants evaluated the program pre-feedback and post-feedback using the rubric. After viewing both versions’ decision-flow diagrams, they sketched their ground-truth decision flow and compared it with the system’s baseline and post-repair diagrams. Sessions concluded with a brief user experience interview (refer to tutorial videos and rubrics in the Supplement).


\begin{table}
    \centering
    \begin{tabular}{|c|c|c|c|}
    \hline
       Participant ID  & Highest Soccer League Played & Years of Coaching Experience & Gender \\
       \hline
       1 & Competitive & 1 & Male \\
       2 & Semi-Professional & 0 & Female\\
       3 & Competitive & 1 & Male\\
       4 & Competitive & 0 & Male\\
       5 & Competitive & 0 & Female\\
       6 & Semi-Professional & 1 & Male\\
       7 & Semi-Professional & 2 & Female\\
       8 & Semi-Professional & 0.5 & Male\\
       9 & Competitive & 2 & Female\\
       10 & Competitive & 0.5 & Female\\
       11 & Competitive & 0.5 & Male\\
       12 & Semi-Professional & 7 & Male\\
       13 & Competitive & 0 & Female\\
       14 & Semi-Professional & 3 & Male\\
       15 & Competitive & 0 & Male\\
       16 & Semi-Professional & 3 & Female\\
       17 & Competitive & 1 & Female\\
       18 & Semi-Professional & 0 & Male\\
       19 & Semi-Professional & 3 & Female\\
       20 & Semi-Professional & 0 & Male\\
    \hline
    \end{tabular}
    \caption{The summary of the participant backgrounds. The ``competitive'' league includes soccer club and competitive-level intramural leagues. The ``semi-professional'' league includes varsity or higher level. The ``recreational'' league includes recreational intramural leagues.}
    \label{tab:participants}
\end{table}

\section{Results}
\subsection{Metrics for Quantifying the Alignment of Learned Program Behaviors with User Intent}
We quantify the alignment in two dimensions. First, the correctness of the program behavior is evaluated 

\noindent\textbf{Correctness.}  
This metric evaluates the program execution (rollouts). Users observe three narrated demonstrations of the program and scored them based on the rubric that we showed them before they provided their initial narrated demonstrations for teaching to ensure that the rubric is reflected in their teaching. Raw rubric scores were normalized to a percentage scale. The correctness score for each program execution $i$ is defined as
\[
C_i = \frac{s_i}{s_{\max}} \times 100,
\]
where $s_i$ is the participant’s rubric score and $s_{\max}$ is the maximum possible rubric score.\\
\textbf{Completeness.}  
This metric complements correctness. The program executions that users evaluated may not cover all branching code paths created by different conditional statements. Thus, completeness quantifies the alignment of the overall program structure (visualized as decision flow diagram) with the user intent. Recall, in the study, each user sketched their ground truth decision flow and compared it with the system's~\ref{sec:study_procedure}. Let $G_{sys} = (V_{sys}, E_{sys})$ denote the set of nodes and edges in the system-generated decision flow, and $G_{gt} = (V_{gt}, E_{gt})$ denote the ground-truth's. Completeness is calculated as the percentage of ground-truth elements (nodes and edges) that also appear in the system’s flow:
\[
\text{Completeness} = \frac{|V_{sys} \cap V_{gt}| + |E_{sys} \cap E_{gt}|}{|V_{gt}| + |E_{gt}|} \times 100.
\]


\subsection{Statistical Analysis Method}
Wilcoxon signed-rank test~\cite{Woolson2007} was used to evaluate to conduct one-sided hypothesis test on whether alignment improved post-feedback within subjects. Wilcoxon test was used instead of paired t-test because participants were sampled from non-Gaussian distribution, with particular set of skills (e.g. playing in competitive or semi-professional soccer leagues). The \texttt{scipy.stats.wilcoxon} function from the SciPy library was used. 


\begin{figure}
    \centering
    \includegraphics[width=0.7\linewidth]{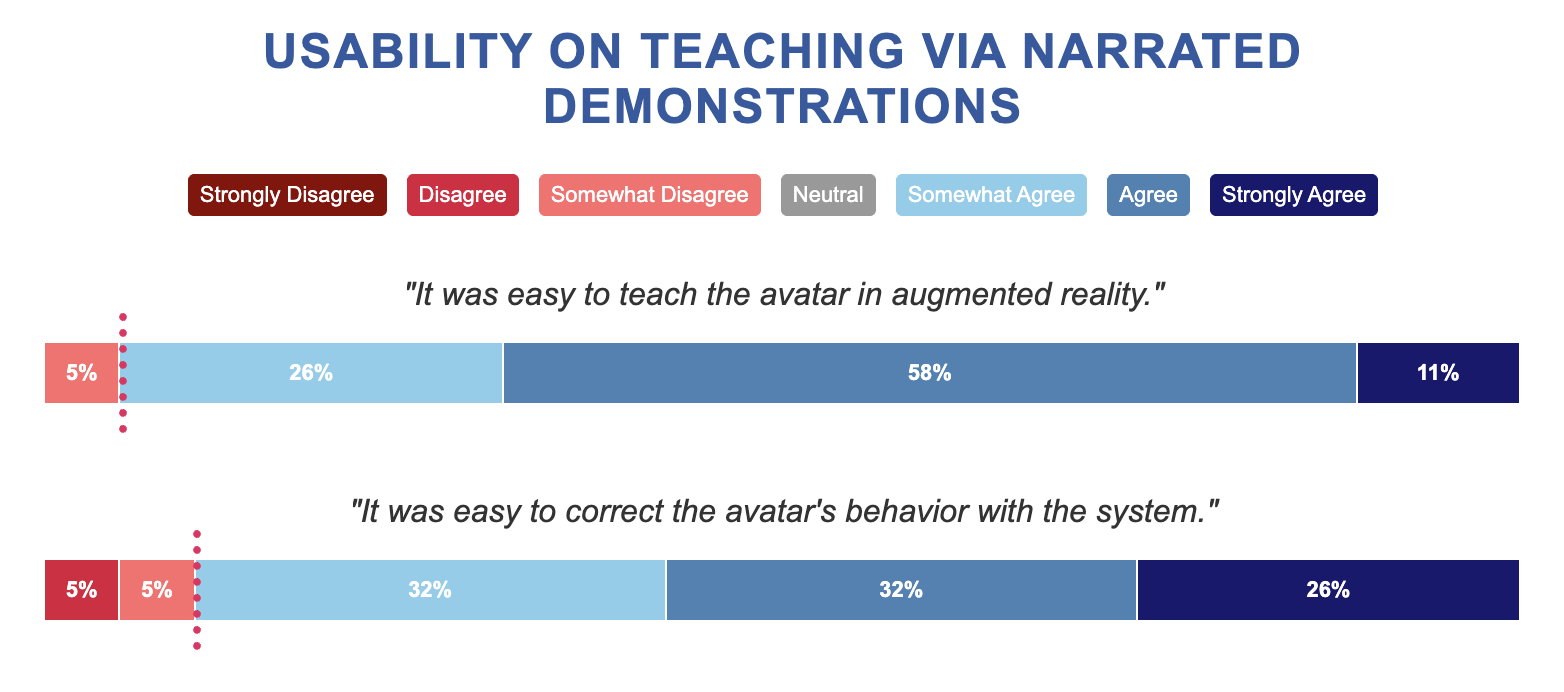}
    \caption{The distributions of user scores related to each statement is shown. }
    \Description{Distributions of user scores for each statement, shown as a bar plot.}
    \label{fig:likert}
\end{figure}

\subsection{User Experience on Teaching via Embodied Narrated Demonstrations (RQ1)} Users authored programs modeling diversity of tactics per scenario (refer to Supplement~\ref{supp:program_diversity}). 95\% of the users agreed that it was easy to teach soccer tactics via narrated demonstrations (Fig.~\ref{fig:likert}). However, users also pointed out some limitations of the teaching interface, where two themes emerged. First, 8 out of 20 participants raised issues regarding the pre-scripted environment agents' behaviors not aligning with their expectations to provide realistic teaching. The pointed out unrealistic movement (linear/stiff paths, stationary or slow defenders) and timing/latency that diverged from typical live play. For example, one noted “Players can only run in linear paths… addition of curved paths would be more realistic” (P2). Another shared “I didn’t feel that I was actually under pressure… unrealistic how slow [defenders are] moving” (P15). The other theme related to the lack of real-time feedback from the system to the participants while they were teaching. A few participants wanted two‑way feedback/acknowledgment from avatars during the teaching phase. “There’s no feedback from the [VR] players… felt a little off compared to real life.”(P13) “It felt a little bit like I was talking to the void… [I wanted] some way of having… feedback.” (P18)

\begin{figure}[t]
    \centering
    \includegraphics[width=1\linewidth]{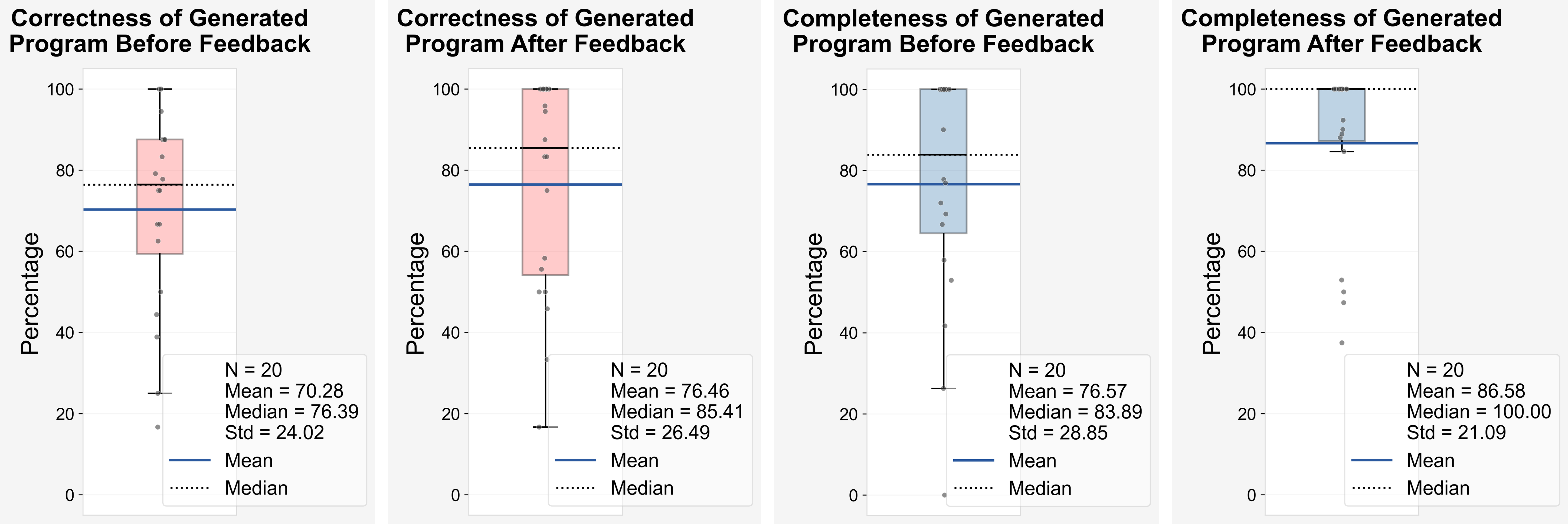}
    \Description{Box plots showing the average and standard deviation of percentage score before and after feedback.}
    \caption{Box plots showing the average and standard deviation of percentage score of the initial synthesis before feedback, and the percentage score of the synthesis after feedback. A score of 100\% on correctness means that the system execution performed all the tasks correctly according to the user, scored from a pre-determined rubric for that scenario. A score of 100\% on completeness means that the decision flow chart outputted by the system is 100\% aligned with the user's own decision flow chart.}
    \label{fig:before-after-completeness-correctness}
\end{figure}


\subsection{Initial System Learning from Narrated Demonstrations of Users (RQ2)}
Pre-feedback alignment of learned programs is shown in Fig.\ref{fig:before-after-completeness-correctness} (first plot from the left) and Fig.\ref{fig:before-after-completeness-correctness} (third plot from the left). The wide variance likely reflects differences in instructional quality. Among the 10 participants whose programs achieved correctness or completeness scores greater than 80\%, their narrations shared two traits: (i) 9/10 explicitly stated the purpose behind actions (e.g., Now your teammate will make an overlapping run, and you want to pass it to them so that they can score the goal.’’ (P5)), and (ii) 7/10 articulated multiple possible outcomes to keep executions correct under varied conditions (e.g., We’re gonna see if the shot is open, I’m gonna take it, but if it’s not, now I’m gonna pass it. Uh, or, let’s say I got covered, I could dribble a bit, and then I can take the shot.’’ (P15)). In contrast, 7 participants whose learned program scored less than 80\% in either metrics exhibited the opposite, either did not explain the rationale (3/7) or did not consider multiple outcomes (4/7).

\subsection{System Learning from User Feedback (RQ3)}
Due to limited study time, participants provided at most two rounds of feedback: one round on the decision-flow diagram and one round on the system’s narrated demonstrations. Participants provided 1.1$\pm$0.4 feedback on average (1 none, 16 one, 3 two). Pre/post alignment results appear in Fig.~\ref{fig:before-after-completeness-correctness}. For improvement analyses, we excluded one participant whose programs scored 100\% pre- and post-feedback since there was no room for improvement. 

\begin{figure}
    \centering
    \includegraphics[width=.75\linewidth]{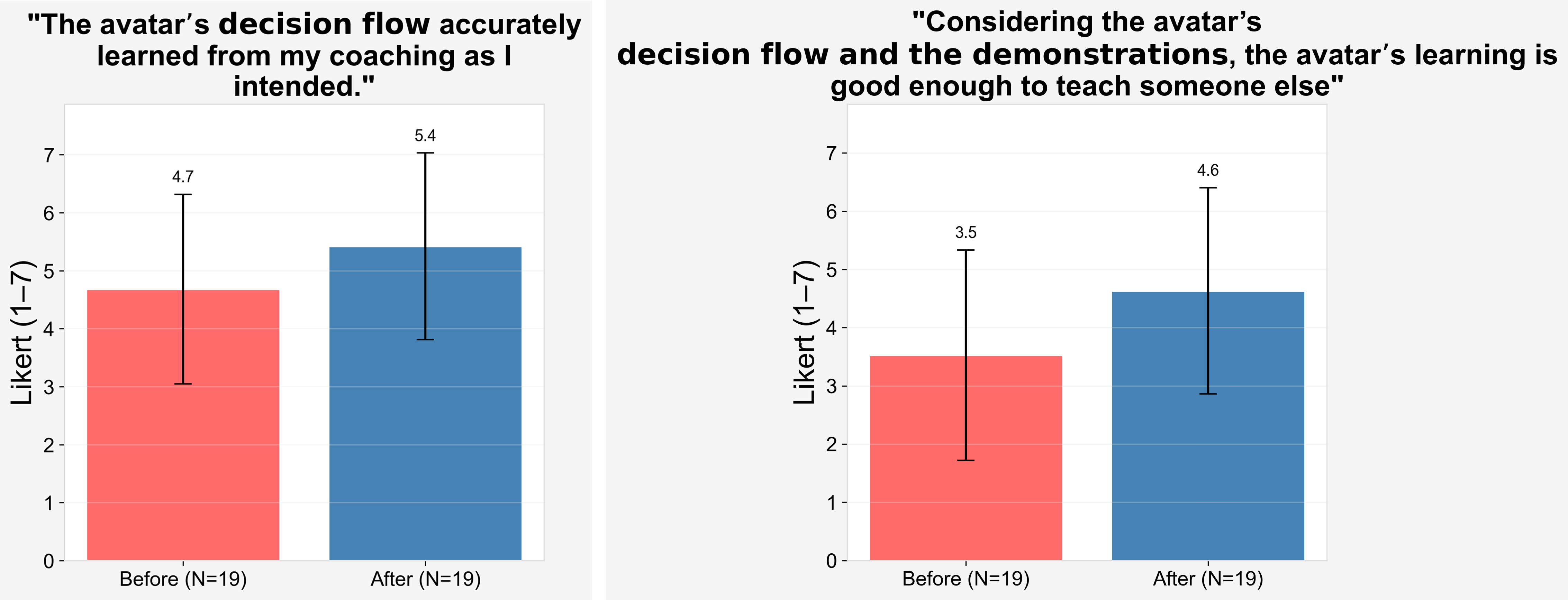}
    \caption{Mean Likert ratings (1–7; higher = more agreement) before vs. after feedback. Left: the avatar’s \textbf{decision flow} accurately reflected what I intended (Before: 4.7, After: 5.4). Right: both the avatar's \textbf{decision flow and demonstrations}, were accurate enough to teach someone else (Before: 3.5, After: 4.6). One participant did not provide feedback and was excluded.}
    \Description{Bar graphs showing Likert scale score improvements before and after feedback.}
    \label{fig:likert-improvement}
\end{figure}

\noindent\textbf{Feedback improved program alignment.} Completeness improved significantly (N=12; mean=16.68\%; median=18.52\%; std=36.54\%; $p=0.03$), while correctness showed a positive but non-significant trend (N=18; mean=6.87\%; median=9.73\%; std=22.42\%; $p=0.10$). To capture qualitative gains not reflected in these alignment metrics, participants also provided 7-point Likert ratings (Fig.~\ref{fig:likert-improvement}), which showed significant improvements in perceived correctness of the decision flow ($p=0.03$) and in overall alignment between the system’s decision flow plus narrated demonstrations and user intent ($p=0.01$).

\noindent\textbf{Failure cases.} Seven of 20 participants did not improve on either correctness or completeness after feedback, due to three sources:
\begin{enumerate}
\item \textbf{Deadlocks (4/7):} The synthesized program deadlocked because a learned condition was never satisfied. The system did not explain which condition blocked execution, preventing users from effectively debugging the issue.
\item \textbf{Insufficient APIs (2/7):} Two participants attempted to correct the curvature of the avatar’s movement trajectories, but our APIs supported only optimal path planning/control and could not accommodate these adjustments.
\item \textbf{Context loss during repair (2/7):} To reduce synthesis time, we omitted original narrated videos from the LLM repair prompt when participants provided feedback on the system’s narrated demonstrations. In two cases, participants gave feedback that implicitly relied on earlier teaching (assuming the system retained that context), which led the LLM to delete intended branches. For example, P17 ended feedback with “Okay, we can stop here,” which the LLM took literally and produced a truncated program.
\end{enumerate}




\subsection{User Experience on Program Editing (RQ4)} 
While correcting system learning, 16/20 participants found the decision flow easy to understand and debug: “Felt like… if-then… I liked that the decision flow was… logically laid out.” (P7) “Correcting the decision flow was pretty smooth… not challenging.” (P14) However, 4/20 reported difficulty with comprehension, noting confusion distinguishing arrows, edges, and nodes: ‘‘Understanding what the arrows were versus the… edges versus the nodes was a little confusing in the beginning’’ (P5). 

All 20/20 participants found the system’s narrated demonstrations easy to follow and diagnose in general: “Very intuitive to make annotations and identify where there were errors” (P1). “It was easy to understand what the avatar got from me based on the narrations and demonstrations” (P8). However, one participant cited unclear terminology was used in one of the system narrations: “Language… ‘recycle the attack’… a little bit unclear” (P18). Regarding UI, while 17/20 participants reported that they were able to sufficiently express their feedback, 3 participants requested richer feedback tools for specifying trajectories: “It would help to be able to \textit{draw} trajectories… more sophisticated motion” (P10).

\section{Discussion}
In this section, we present our analysis of the user study and discuss the limitations of our system.

\subsection{Learning Collaborative Physical Activities Needs Iterative Corrections}
Our findings show that programs generated from users’ initial narrated demonstrations were often unreliable as models of collaborative physical activities (average correctness: 70.28$\pm$24.02\%; average completeness: 76.57$\pm$28.85\%). Performance also varied substantially across participants (SD: 24.02\% for correctness; 28.85\% for completeness), suggesting that one-shot learning from a single batch of narrated demonstrations frequently failed to capture key task intent.

\textbf{Model misinterpreted intent from demonstrations.} We observed recurring VLM misinterpretations of users’ intent: the vision-language model (VLM) generally captured narrated content, but often failed to infer intent from demonstration videos across the three tactical scenarios. For example, in the lure scenario (also used as the running example in Sec.~\ref{sec:system_architecture}), a user moved to the side of the field while staying near a teammate with the ball and narrated, “to lure the opponent to the side of the field, move around here.” Although the demonstrations implicitly conveyed an additional constraint—to remain close to the teammate—the VLM extracted only “move to the side of the field,” allowing positions anywhere on that side, including near the goal far from the teammate.

\textbf{Fluctuations in teaching quality strongly shaped outcomes.} We also observed a wide spectrum in users’ teaching quality, which likely contributed to the large variability in initial program accuracy. Among the 10 participants whose programs achieved correctness or completeness scores greater than 80\%, narrations commonly exhibited two traits: (i) 9/10 explicitly stated the purpose behind actions (e.g., “Now your teammate will make an overlapping run, and you want to pass it to them so that they can score the goal.” (P5)), and (ii) 7/10 articulated multiple possible outcomes to keep executions correct under varied conditions (e.g., “We’re gonna see if the shot is open, I’m gonna take it, but if it’s not, now I’m gonna pass it. Uh, or, let’s say I got covered, I could dribble a bit, and then I can take the shot.” (P15)). In contrast, among the other 10 participants whose learned programs scored less than 80\% on either metric, 7 participants exhibited the opposite pattern—either not explaining the rationale behind actions (3/7) or not accounting for multiple outcomes (4/7).

\subsection{On the Effectiveness of the Interfaces to Teach, Inspect, and Correct System Learning}

Given the limitations of both the generative model and users’ teaching, we now examine how our interfaces help users teach, inspect, and correct the system’s learning. Our system architecture provides a unified interaction modality: narrated demonstrations serve as the modality for teaching, for the system’s explanations of its learning, and for users’ corrective feedback. Participants generally perceived this modality as natural and easy to use. As shown in Fig.~\ref{fig:likert}, 95\% (19/20) of users reported that it was easy to teach collaborative activities to our system. 80\% (16/20) found the decision flow easy to understand, and all participants reached a consensus that the system's narrated demonstrations were easy to understand. Finally, 90\% (18/20) indicated that it was easy to correct the system’s learning.

Beyond perceived usability, users were able to leverage these interfaces to correct the system’s learning and help it converge toward more accurate models of collaborative physical activities. Even with only a single round of feedback for most participants (16/20, due to time constraints), program completeness increased by 16.68\% on average (SD = 18.52, $p = 0.03$). Correctness also showed a positive, though non-significant, gain of 6.87\% (SD = 22.42, $p = 0.10$). These quantitative improvements were mirrored in participants’ subjective ratings of their ability to steer the modeled programs toward the intended soccer tactics (Fig.~\ref{fig:likert-improvement}). In interviews, several participants further noted that additional iterative rounds of feedback would likely have yielded even more accurate program representations.

\subsection{Limitations}
Our approach currently lacks mechanisms for detecting inconsistencies, ambiguities, or under-specification in users’ instructions to better elicit their domain knowledge. The system takes narrated demonstrations at face value when generating or repairing programs. As a result, the system amplified individual differences in teaching ability, leading to wide variation in the accuracy of its learned models. Furthermore, the observed issues with user inspection raise questions about the scalability of our system. Twenty-five percent (5/20) of participants reported that the decision-flow diagrams became difficult to interpret as they grew in complexity, often containing redundant nodes and edges. This problem is likely to worsen as the number of agents or the complexity of activities increases. One potential workaround is to hide the decision-flow diagrams and rely solely on system executions for feedback. However, this would require users to watch too many executions to cover all branching code paths induced by contingencies, underscoring the need for mechanisms that support holistic inspection of program structure without cognitively overloading users to solicit feedback.

Although users’ feedback improved the average correctness of system learning, it also increased fluctuations in correctness outcomes after feedback, i.e., standard deviation increased (Fig.~\ref{fig:before-after-completeness-correctness}, left). Beyond individual variation in instruction quality, our analysis points to two major sources of error. First, the expressiveness of the API library was a major constraint. While the APIs enabled the insertion of domain knowledge to facilitate learning from only a few narrated demonstrations, they also restricted the spectrum of behaviors that could be modeled. For example, one user attempted to teach an avatar to follow a curved trajectory, which our APIs did not model; as a result, the avatar moved in a straight line instead. Second, our inspection interface did not support handling deadlocks in program execution. In 10\% (2/20) of user evaluations, programs halted during user inspection phase because none of their preconditions were satisfied. The system could narrate the program logic but lacked mechanisms to detect and communicate to users that its halted because certain preconditions are not satisfied.

\section{Conclusion \& Future Work}
Modeling collaborative physical activities has broad implications for HCI, from behavior modeling to tutorial and assistance systems. This makes the design of interaction techniques for iteratively correcting models of collaborative activities an important area for HCI research. In this work, we present an interactive system that enables users with no programming experience to author programs that model collaborative activities as intended. The physical nature of these activities requires interfaces that not only support teaching them, but also enable users to inspect and correct the system’s learning. We therefore propose interfaces that use narrated demonstrations as a unified, bidirectional modality: users teach and correct the system through paired physical demonstrations and natural language, and the system communicates its learned behavior through the same modality, without exposing raw code. Our user study in a tactical soccer setting, together with an existence proof in a factory environment, demonstrates the feasibility of this approach. Taken together, our findings point toward interactive task-learning systems that leverage interpretable program representations and human-in-the-loop refinement to capture collaborative behaviors from user instructions.

Our study surfaces key design insights that future systems should account for. It shows that programmatically embedding domain knowledge via APIs can enable learning complex collaborative activities from only a handful of narrated demonstrations, despite the difficulty of this problem. However, these APIs must be carefully designed through in-depth analysis of user needs and practices in each target domain to support a broad spectrum of users. Moreover, while modeling collaborative activities as programs makes it easy to insert domain knowledge, it can also introduce deadlocks. To help users localize and correct such situations, underlying generative models could be further instructed to create code routines for detecting deadlocks and explicitly articulating the sources of these errors to users for correction.

The study results also raise open questions for the HCI community. One concerns scalability, particularly in relation to decision-flow diagrams. Although these diagrams are useful for holistically visualizing overall program structure, they become increasingly difficult for users to comprehend as the complexity of activities or the number of agents grows. One potential direction is to display hierarchical representations, abstracting subsets of nodes and edges into higher-level units to reduce visual complexity. However, algorithms for determining appropriate levels of abstraction as a function of task complexity remain to be investigated.

Another open question relates to detecting inconsistencies, ambiguities, or incompleteness in user instructions in order to better elicit accurate domain knowledge. It may be fruitful to incorporate formal languages and algorithms from formal methods~\cite{Clarke2018HandbookModelChecking} to mathematically formalize the specifications in users’ instructions and detect logical inconsistencies, ambiguity, or under-specification. At the same time, the expressiveness of existing formal languages for capturing relevant spatial and temporal constraints needs to be examined, along with whether the underlying algorithms can be executed efficiently enough to support real-time inspection and maintain high usability. Another promising direction is to incorporate agentic reasoning, in which LLM-based agents proactively infer missing details and resolve inconsistencies or ambiguities in user instructions based on context~\cite{Plaat2025AgenticLLM}.

\bibliographystyle{ACM-Reference-Format}
\bibliography{sample-base}

\newpage
\appendix

\section{User Study Supplement}
\subsection{Tutorial Video}\label{supp:tutorial} Here is \href{https://drive.google.com/file/d/1Fyzr8lflQ9z49QNBZYxYfmKwWahcZ54M/view?usp=sharing}{the link to the tutorial video} that all participants watched at the beginning of the study. This video covers how to teach, inspect, and correct the system via narrated demonstrations. 

\subsection{User Scenarios}\label{supp:user_scenarios}
In the study, each participant was assigned one of three pre-scripted scenarios. For each scenario, we prepared a tutorial video explaining the behaviors of the teammate(s) and opponent(s) as well as the tactical goals to be achieved. \href{https://drive.google.com/drive/folders/1lW794wGa4ZWCNyqECjHYQu4ajLQN9MIH?usp=drive_link}{The shared drive} contains the three tutorial videos. The scenarios were: (1) lure, (2) overlap, and (3) distribute.

\begin{figure}
    \centering
    \includegraphics[width=\linewidth]{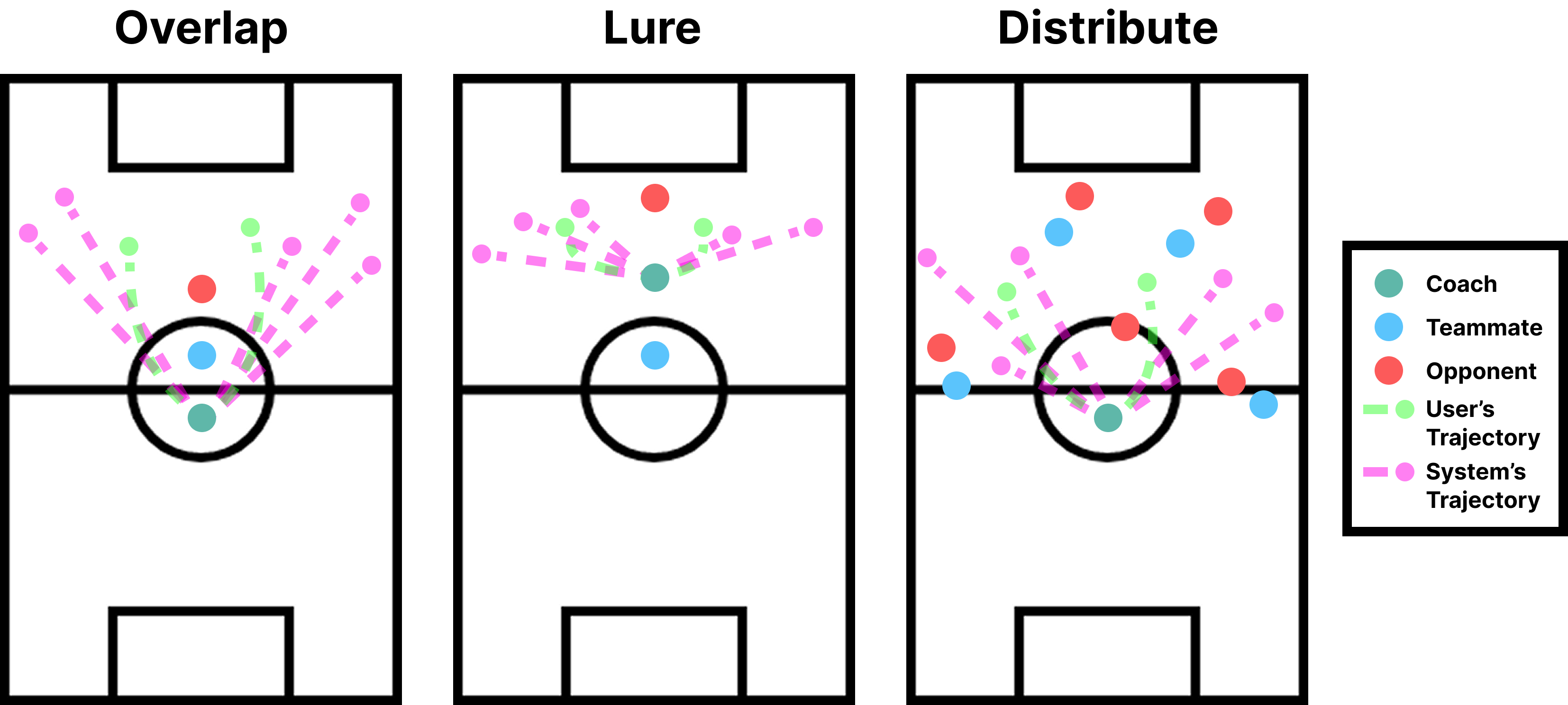}
    \caption{Three plots overlay participants' demonstration trajectories with execution trajectories from the learned probabilistic programs, each corresponding to one of the three user scenarios in Appendix~\ref{supp:user_scenarios}.}
    \label{fig:generalizability}
\end{figure}

\subsection{Rubric for Each Scenario}\label{supp:rubric}
Here is \href{https://docs.google.com/document/d/1sMUKFHhMpWTyw3S-Vz2vEumMAXWIH6npNlNorEJzCK4/edit?usp=sharing}{the documentation of the three rubrics for the corresponding scenarios}, which participants used to score the correctness of learned programs after watching their executions. Each participant was shown the rubric for their assigned scenario before teaching the system to ensure instruction that could yield full scores. Each rubric consists of multiple statements, each rated 0, 1, or 2: a score of 2 indicated the program completed the statement exactly as intended, 1 indicated partial or approximate completion, and 0 indicated failure. Participants watched three program executions and scored each with the rubric; their reported score was the average of the three. These results are shown in Fig.~\ref{fig:before-after-completeness-correctness} (left two plots).

\subsection{User Interview Questionnaire} 
\begin{enumerate}
  \item How was your experience teaching in augmented reality?
  \begin{enumerate}
    \item What did you find intuitive?
    \item What did you find challenging?
    \item How can we improve the system?
  \end{enumerate}

  \item How was your experience correcting the avatar's behavior by commenting on the avatar's decision flow?
  \begin{enumerate}
    \item What did you find intuitive?
      \begin{enumerate}
        \item Was it easy to understand the decision flow?
      \end{enumerate}
    \item What did you find challenging?
    \item How can we improve the system?
  \end{enumerate}

  \item How was your experience correcting the avatar's behavior by \emph{directly annotating and commenting} on the avatar's narrations and demonstrations on the soccer field?
  \begin{enumerate}
    \item What did you find intuitive?
      \begin{enumerate}
        \item Was it easy to understand what the avatar learned based on its narrations and demonstrations?
      \end{enumerate}
    \item What did you find challenging?
    \item How can we improve the system?
  \end{enumerate}

  \item How was the avatar's learning compared to what you intended to teach?
  \begin{enumerate}
    \item How would you think of the avatar teaching someone else based on what it learned from you?
      \begin{enumerate}
        \item What needs to be improved?
      \end{enumerate}
  \end{enumerate}
\end{enumerate}

\subsection{Generalizability of Learned Programs}
The learned Scenic probabilistic programs did not simply replay participants' narrated demonstrations. In the study, each participant provided two demonstrations of the same task, and these variations were modeled as distributions in Scenic programs. Semantically, a Scenic execution samples a concrete scenario from the distribution it encodes and generates it in simulation. Fig.~\ref{fig:generalizability} overlays two demonstration trajectories of a participant with five program execution trajectories for each of the three scenarios: overlap (left), lure (middle), and distribute (right), corresponding to Participants 9, 10, and 11. These participants produced learned programs that achieved 100\% correctness and completeness via interactive program generation. 

\subsection{Generalizability of Teaching Methodology}
The proposed mechanism for users to teach physical activities and tasks can generalize to multiple teaching modalities. Beyond the embodied agent interaction presented in the paper, it can also support user's teleoperation of an agent (e.g., a robot) either in reality or virtual reality as well as conventional lead-through demonstration in robotics, where a user physically guides the robot by moving its joints, gripper, and its any other components while narrating the task. 

\subsection{Diversity of User-Authored Programs}\label{supp:program_diversity}
For each scenario, a rubric with three to four task requirements were provided (Appendix~\ref{supp:rubric}), but these were potentially many ways to complete them. As a result, users authored programs reflecting diverse strategies per scenario. For example, in the lure scenario, some participants used fixed spatial thresholds (e.g., specific distances or lanes) while others encoded more conditional, temporal logic (e.g., “if the defender steps, then pass; otherwise dribble or reset”). In the overlap scenario, some prioritized getting wide first, another on dribbling towards goal immediately, and others focused on the exact shape and timing of the run, leading to different code branching structures and termination conditions. In the distribute scenario, some users first scanned the field to identify open teammate, while others moved around to create open lanes for a teammate. Together, these variations show that our technique supported users to externalize diverse strategies reflecting their own domain knowledge. 

\subsection{System Implementation}\label{appendix:system_implementation}
\subsubsection{Teaching, Inspection, Editing Interfaces} The details of our three interfaces are explained in our tutorial video linked in Appendix~\ref{supp:tutorial}. 

\subsubsection{APIs} 
\begin{figure}[!htbp]
    \centering
    \includegraphics[width=\linewidth]{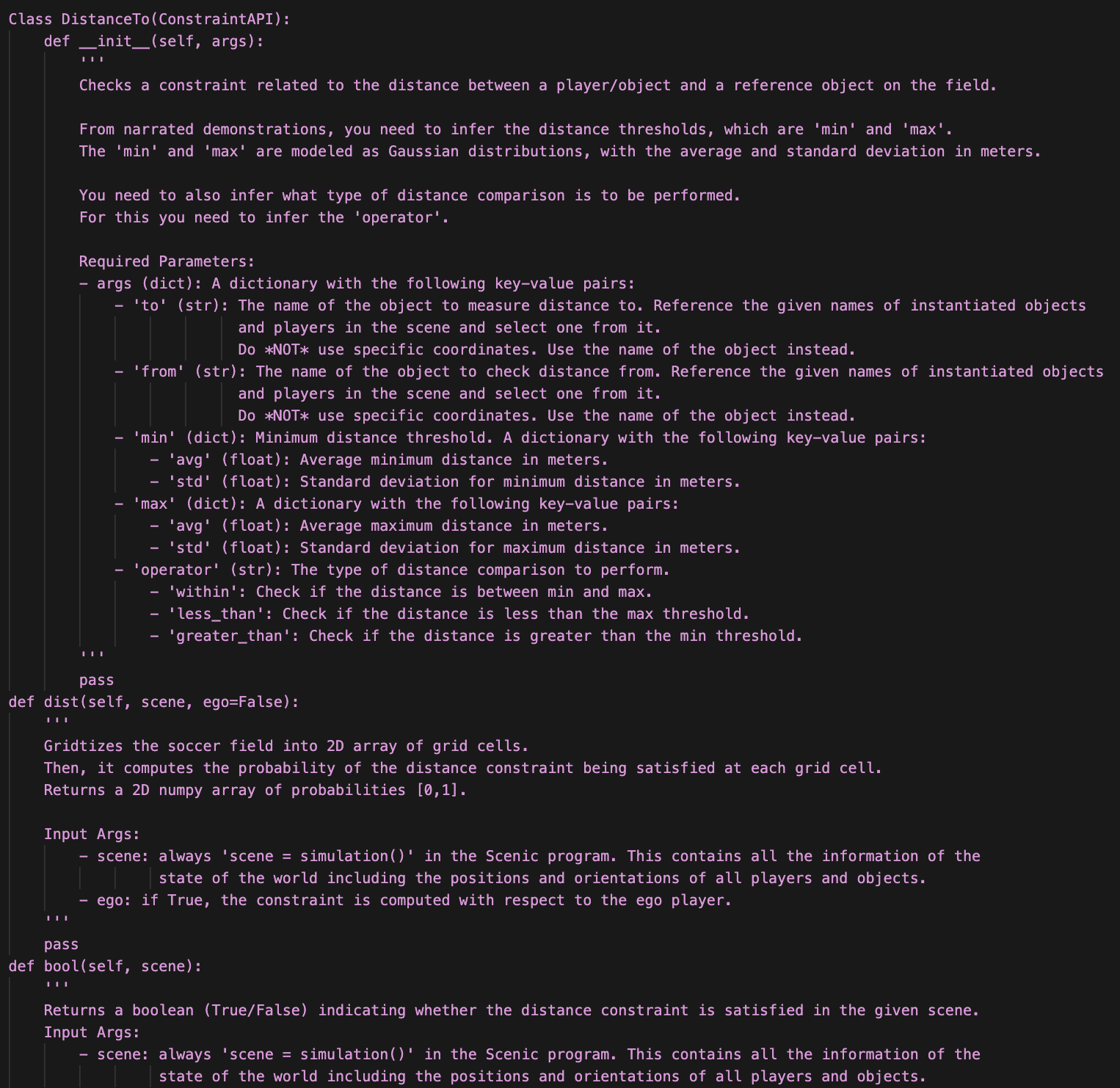}
    \caption{An annotation of an API we provided to a large vision language model.}
    \label{fig:api}
\end{figure}

Our study personnels consisted of soccer players with the history of playing in the competitive soccer leagues. We used our domain knowledge of soccer to model the sets of action space and the physical constraints as APIs. For action space, we provided APIs that execute moving/dribbling, passing, shooting a goal, and triggering a teammate to pass the ball to the user. For modeling the physical constraints, we APIs that can validate conditions such as (1) ball possession, (2) distance relations, (3) horizontal/vertical relations, and (4) existence of passing lane between two players of interest. To validate these conditions, the APIs were internally given access to the world states (e.g. positions, orientations, ball possession of each player) in mixed reality simulation at each timestep. The annotations of all APIs were provided to the large vision language model (LVLM) for program generation. For example, Fig.~\ref{fig:api} shows the annotation of \texttt{DistanceTo()} API which checks for the distance relation between two players of interest.

\subsubsection{System Prompt} 
\begin{figure}[!htbp]
    \centering
    \includegraphics[width=\linewidth]{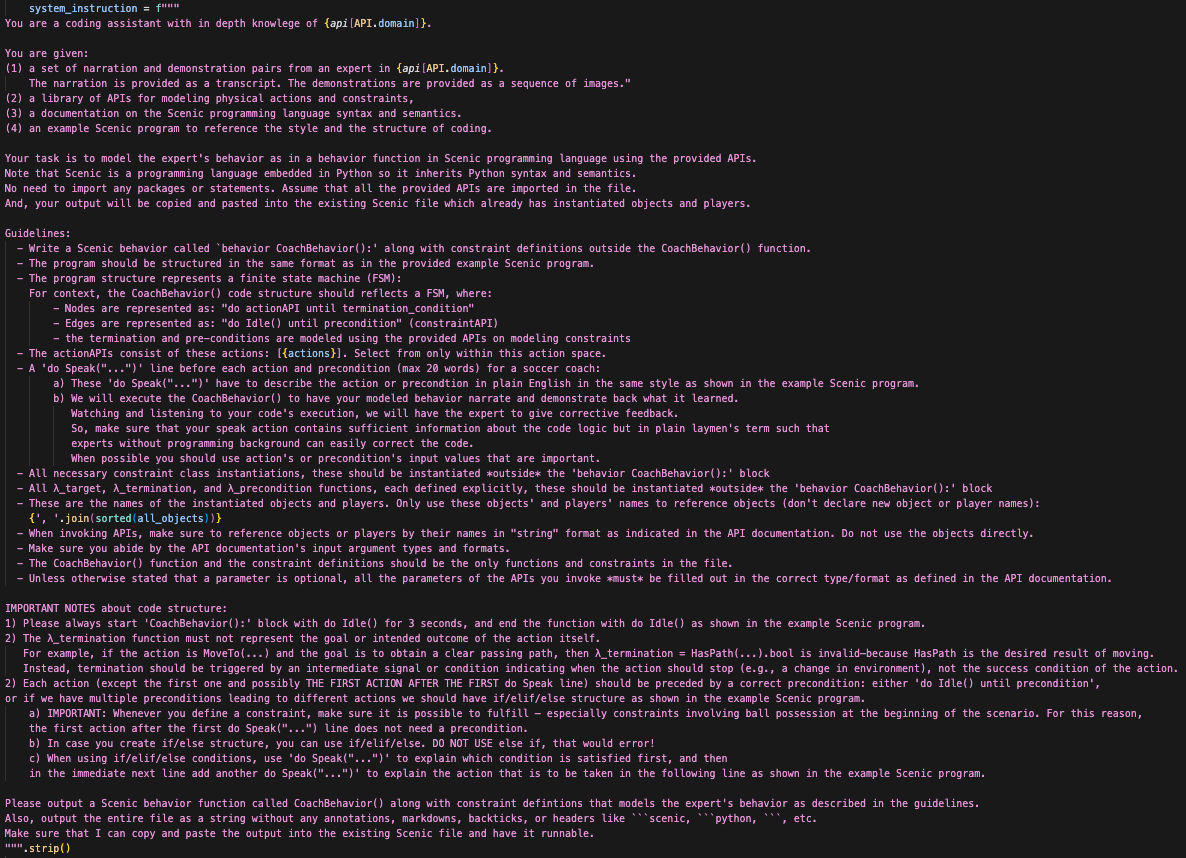}
    \caption{The system prompt used in the user study for instructing LVLM how to code.}
    \label{fig:system_prompt}
\end{figure}

The system prompt used to instruct LVLM generate programs for our user study is shown in Fig.~\ref{fig:system_prompt}.

\FloatBarrier

\section{Existence Proof in Factory Environment}\label{supplement:factory}

\subsection{Factory Tasks and Scenarios}

In the factory setting, participants taught collaborative box-packaging interactions between a mobile robot and a human worker. Unlike the soccer study, the scenarios were \emph{not} pre-scripted: before teaching, each participant authored the initial scene from scratch by placing the human, robot, boxes, and other objects anywhere they wished in the Unity environment. Each participant was assigned one of two scenarios designed to exercise complementary sides of the collaboration rubric. In the \textbf{robot-focused scenario} (robot as learner), the participant's demonstrations focused on teaching the \emph{robot} how to collaborate with the human worker, by fetching and delivering boxes to support the human's packaging workflow. In the \textbf{human-focused scenario} (human as learner), the participant's demonstrations focused on teaching the \emph{human} how to collaborate with the robot, by deciding when and how to signal the robot to bring a new box and then packaging the box once it arrived.

\subsection{Study Protocol}
The existence proof followed the same study procedure as the main soccer study (Appendix A), with modifications to support authoring-from-scratch in the factory setting. We recruited three participants for the factory existence proof following the same recruitment protocol as in the study~\ref{sec:participants}, without imposing any inclusion criteria. Before teaching, we provided a short live tutorial demonstrating the new authoring-from-scratch interface: how to place agents and objects in the scene and what actions the agents could take, how to start/stop recording demonstrations, how to provide feedback to the decision flow diagram and system execution of the generated program, and how to use the rubric to evaluate behavior.

\subsection{Rubrics for Factory Scenarios}

As in the main study, participants used the rubric associated with their scenario to judge whether the system's learned behavior matched their intent.

\paragraph{Robot-focused scenario rubric (teaching the robot).}
\begin{enumerate}
    \item Robot fetches the box.
    \item Robot moves to the correct position next to the human to deliver the box.
    \item Robot puts down the box.
    \item Human is able to package the box.
\end{enumerate}

\paragraph{Human-focused scenario rubric (teaching the human).}
\begin{enumerate}
    \item Human signals for the robot to bring a new box.
    \item Robot brings box to human.
    \item Human packages the new box.
\end{enumerate}

As in the soccer setting, each rubric item was rated on a 0--2 scale (0 = not achieved, 1 = partially achieved, 2 = fully achieved).

\subsection{System Implementation for the Factory Setting}

\subsubsection{Action and Constraint APIs}

The factory setting extends the domain-specific API library described in the main text with actions and constraints tailored to human--robot box-packaging. Concretely, we provide action APIs such as moving to a target position, picking up an object, putting down an object, packaging a box, and raising a hand. In addition to the spatial and logical constraints used in the soccer setting (e.g., distances, regions, and compositional logic), the factory environment includes constraints that reason about object and agent state, such as whether a box is currently held or already packaged and whether a particular agent (e.g., the human) has raised their hand to request assistance. These APIs allow the system to express and infer conditions such as ``the human has requested a new box'', ``the robot is holding a box'', or ``the box has been fully packaged''.

\subsubsection{Authoring from Scratch via LLM-Generated Scenes}

A key difference from the soccer setting is the authoring-from-scratch workflow on the desktop instead of demonstrating on pre-scripted scenarios in the mixed reality environment. In the factory setting, each time the participant presses the \textbf{Record} button to speak to author a new scene and demonstration, the system:
\begin{enumerate}
    \item Sends a prompt to a language model that includes:
    \begin{itemize}
        \item the current list of available prefabs (e.g., human, robot, box),
        \item the catalog of available actions (as above),
        \item any clicked positions or selected objects from the interface.
    \end{itemize}
    \item The LLM responds with a JSON specification that either:
    \begin{itemize}
        \item spawns new objects in the scene (using prefabs and world positions), or
        \item updates existing objects by assigning them actions to execute.
    \end{itemize}
\end{enumerate}

At this stage, the LLM's output provides a 1-to-1 scene configuration for the demonstration; it does not yet synthesize a Scenic probabilistic program. After the participant has provided multiple demonstrations of the same scenario, we then apply our program synthesis pipeline (described in the main text) to generate a Scenic probabilistic program that generalizes across demonstrations.

\subsubsection{Prompt for Scenario Authoring}

For completeness, we share below the prompt that we used for LLM-based scenario authoring in the factory setting. In the actual system, \texttt{prefabListJson} and \texttt{actionCatalogJson} are filled with JSON describing the available prefabs and actions.

\begin{verbatim}
You are a scenario planner for a Unity simulation.

Output ONLY raw, minified JSON that matches this schema EXACTLY
(no markdown, no commentary):

{
  "objects": [
    {
      "prefab": "string",
      "name": "string (REQUIRED)",
      "position": [x, y, z],
      "rotation": [x, y, z],
      "scale": [x, y, z],
      "actions": [
        {
          "name": "string",
          "parameters": { "paramName": "value" }
        }
      ]
    }
  ]
}

Rules:
- Always return an "objects" array (it can be empty).
- Every object MUST have a unique "name" field.
- Never include explanations.

SPAWNING NEW OBJECTS:
- Include: "prefab", "name", "position", "rotation", "scale".
- Use clicked_positions for placement.

UPDATING EXISTING OBJECTS:
- Include ONLY: "name" and "actions".
- DO NOT include: "prefab", "position", "rotation", or "scale".
- If [selected_objects] contains the object name, it EXISTS —
  do not spawn it again.

Additional rules:
(1) If the user message includes [clicked_positions: ...] and/or
    [selected_objects: ...], treat them as HINTS to resolve deictic
    references like "this one" or "here".
(2) When assigning behaviors to already-spawned objects, identify them
    by a stable name from selected_objects when available.

AllowedPrefabs = {prefabListJson}
Rules:
- You can ONLY spawn objects using these prefab keys.
- Use these exact strings (case-insensitive) for the "prefab" field.
- Do not invent or use any prefab names not in this list.

AllowedActionCatalog = {actionCatalogJson}
Rules:
- The only actions that players/robots can take are provided here,
  based on the entire list of action functions. Objects such as boxes
  CANNOT take actions; only players/robots can.
- Use only "name" values in this catalog for the "actions[].name" field.
- Map Vector3 as {"x": float, "y": float, "z": float}.
- Do not invent functions not listed here.
\end{verbatim}

\end{document}